\documentclass[a4paper,fleqn,longmktitle]{cas-dc}
\ExplSyntaxOn
\let\vbox_unpack_clear:N \vbox_unpack_drop:N
\ExplSyntaxOff

\ExplSyntaxOn
\cs_set:Npn \__first_footerline:
{
  \group_begin:
  \small \sffamily
  \ifnum\theblind>0\relax \else \__short_authors: :~ \fi
  { \rmfamily \itshape Preprint }
  \group_end:
}
\ExplSyntaxOff

\usepackage[authoryear,longnamesfirst]{natbib}
\usepackage{amsmath,amssymb,amsfonts}
\usepackage{underscore}
\usepackage{graphicx}
\usepackage{booktabs}
\usepackage{multirow}
\usepackage{adjustbox}
\usepackage{array}
\usepackage{url}
\usepackage{xcolor}
\usepackage{siunitx}
\usepackage{enumitem}

\newcommand{\PairedHeadline}{All 15 cells reject $H_0: \Delta = 0$ at BH-FDR $q \leq 0.05$ (raw $p \leq 0.0013$; BH-adjusted $q \leq 0.0013$); per-cell $\Delta$ ranges $+0.119$ to $+0.387$.}

\begin{document}
\let\WriteBookmarks\relax
\def\floatpagepagefraction{1}
\def\textpagefraction{.001}

\title[mode=title]{Frozen Foundation-Model Embeddings Discard Small-Lesion Signal in Chest Radiography: Implications for Pre-Deployment Evaluation}

\shorttitle{Small-lesion signal loss in CXR foundation models}
\shortauthors{R. Muthyala et~al.}

\author[1]{Raajitha Muthyala}
    \credit{Methodology, Software, Investigation, Data curation, Writing -- original draft}
    
    \author[1]{Zhenan Yin}
    \credit{Software, Investigation, Visualization}
    
    \author[1]{Alekhya Jilla}
    \credit{Software, Investigation}
    
    \author[2]{Frank Li}
    \credit{Resources, Validation, Writing -- review \& editing}
    
    \author[2]{Theo Dapamede}
    \credit{Resources, Validation, Writing -- review \& editing}
    
    \author[2]{Bardia Khosravi}
    \credit{Resources, Validation, Writing -- review \& editing}
    
    \author[2]{Mohammadreza Chavoshi}
    \credit{Resources, Validation, Writing -- review \& editing}
    
    \author[2]{Judy Gichoya}
    \credit{Conceptualization, Resources, Supervision, Writing -- review \& editing}
    
    \author[1]{Saptarshi Purkayastha}[orcid=0000-0003-3625-534X]
    \cormark[1]
    \ead{saptpurk@iu.edu}
    \credit{Conceptualization, Methodology, Investigation, Supervision, Project administration, Writing -- review \& editing}
    
    \affiliation[1]{organization={Department of Biomedical Engineering and Informatics, Indiana University},
                    addressline={535 W Michigan St., IT 475J},
                    city={Indianapolis},
                    state={IN},
                    postcode={46202},
                    country={USA}}
    
    \affiliation[2]{organization={Department of Radiology and Imaging Sciences, Emory University},
                    addressline={1364 Clifton Rd NE},
                    city={Atlanta},
                    state={GA},
                    postcode={30322},
                    country={USA}}
    
    \cortext[1]{Corresponding author.}

\begin{abstract}
\noindent Frozen vision-transformer (ViT) foundation-model embeddings increasingly serve as the substrate for downstream chest-radiography (CXR) pipelines, yet where small-scale, low-contrast signal is retained or lost in the frozen forward pass has not been systematically quantified across architectures, pretraining domains, and objectives. We probed five frozen ViTs (RAD-DINO, DINOv2-B/14, DINOv3 ViT-7B, BiomedCLIP, MedSigLIP) and a frozen DINO-pretrained ResNet-50 architectural control across three large CXR cohorts (NIH-CXR14, MIMIC-CXR, Emory-CXR; aggregate pool $n=492{,}724$) and ChestX-Det10 ($n=3{,}543$; 1{,}462 small-lesion bounding boxes across Calcification, Nodule, Mass). Each model was evaluated with a small-scale-perturbation panel and a region-aware bounding-box-stratified probe on real lesions, comparing three pooling modes from the same forward pass: classification token (CLS), patch-mean (mean over all final-layer patch tokens), and bounding-box-restricted patch-local. On the perturbation panel, CLS embeddings sat at the chance floor (area under the ROC curve [AUC] 0.500--0.524); patch-mean was indistinguishable from CLS on iso-blur and reticular-fine cells but rose with CLS on larger directional-blur footprints, while disease AUC on globally decided tasks ranged 0.642--0.913. Patch-local probes recovered AUC $\approx 1.0$ from the same forward pass (per-model mean improvement $+0.412$ to $+0.488$); the ResNet-50 control reproduced the chance floor. On ChestX-Det10, image-level CLS classification showed within-class small-versus-large stratum gaps up to $+0.243$ AUC; bounding-box-level patch-local pooling on the same forward pass recovered AUC $\geq 0.899$ on every (model $\times$ class) cell. Frozen ViT embeddings silently suppress small-scale signal at the global-aggregation step; the signal is recoverable from patch tokens conditional on a region of interest.
\end{abstract}




\maketitle

\section{Introduction}
\label{sec:introduction}

Advances in artificial intelligence have led to radiology-focused algorithms designed to aid the interpretation of medical images \citep{li2025embeddings}. Foundation models are large self-supervised vision encoders that emit a single fixed-dimensional embedding per image, which have emerged as the dominant paradigm \citep{fang2025large,dantonoli2025foundation}, and downstream clinical pipelines increasingly treat that embedding as the substrate from which all task-specific decisions are made \citep{khoiwal2024embeddings}. Models such as RAD-DINO, trained on radiology datasets \citep{perezgarcia2025raddino,zedda2025radiodino}, DINOv2 and DINOv3, trained on web-scale natural images \citep{oquab2024dinov2,simeoni2025dinov3}, BiomedCLIP, trained on biomedical image--text pairs \citep{zhang2023biomedclip}, and MedSigLIP, trained on medical image--text pairs \citep{sellergren2025medgemma}, exemplify this paradigm and have been benchmarked extensively on chest X-ray disease classification, including cardiomegaly, pneumonia, and lung-lesion detection \citep{codella2024medimageinsight,liu2025dinov3med}.

Many of the most clinically consequential chest radiography findings, however, are not decided by global anatomy. Early-stage primary lung cancer can present as a 4-8 mm primary nodule \citep{henschke2006survival}; subtle interstitial fibrosis manifests as fine reticular textures with sub-percent contrast against parenchyma \citep{hansell2008fleischner}; ground-glass opacities at the threshold of clinical significance occupy small focal regions; the analogous problem in mammography is the detection of microcalcifications, a textbook small-feature task in which the diagnostic signal is at the limit of pixel resolution. What unites these tasks is a shared spatial-frequency signature: the diagnostic signal is small (well under one percent of the image area), low-contrast (often within $\pm 10\%$ of local parenchymal intensity), and spatially restricted to a single anatomical region. A foundation model whose embedding silently discards information at this spatial scale will, by construction, fail to encode the diagnostic signal that defines these tasks, and crucially, this representational failure cannot be detected by evaluating the same model on disease tasks decided by global anatomy, where the diagnostic signal lies elsewhere in the spatial-frequency spectrum. Pre-deployment evaluation that relies solely on disease-AUC benchmarks is precisely the regime in which silent representational deficits go undetected, alongside the shortcut-learning failure modes documented for radiology AI more broadly \citep{banerjee2023shortcuts}.

This study tests whether such silent small-scale-signal loss occurs in modern ViT foundation models for chest radiography, and where in the model it occurs. We argue that the question is empirically tractable: a panel of small, controlled perturbations injected into clean images functions as a \emph{probe} of the embedding's small-scale capacity. If a frozen embedding cannot linearly discriminate clean images from images perturbed at a given spatial scale, it has, at minimum, discarded enough small-scale signal at that scale to be unable to recover binary class membership, even with optimal pixel-level localization. Conversely, if the embedding can discriminate, it has retained enough small-scale signal to potentially support a diagnostic task at that scale. The probe is \emph{necessary, not sufficient}, for small-feature diagnostic capacity, and an injected-perturbation null result is therefore an actionable warning about deployment to small-feature diagnostic tasks.

Prior work in general computer vision has established that ViTs are substantially more robust to image corruptions than convolutional networks, an advantage attributed to the self-attention mechanism's progressive suppression of local noise across layers \citep{naseer2021intriguing,zhou2022understanding} and to the global aggregation properties of the CLS token, which integrates over thousands of patch tokens and dilutes any single locally-corrupted region. Robustness studies universally measure robustness as degradation in classification accuracy under corruption; they ask whether a model's predictions remain correct when the input is perturbed. This is a different question from whether a model's embedding space encodes the perturbation at all. A model could be robust in the accuracy sense while its embedding fully encodes the artifact, or suppressive in the embedding sense while maintaining accurate predictions on tasks whose diagnostic signal lies elsewhere.

\paragraph{Relation to prior work on CLS-vs-patch-token information.} The general observation that CLS pooling and patch features encode different content has been documented in mainstream ViT literature. \citet{darcet2024registers} showed that DINOv2 develops high-norm tokens in low-information patch positions that store global information, degrading dense-task performance, and that adding ``register'' tokens during pretraining cleans the patch-token feature map. \citet{lappe2025decoupling} showed that in DINOv2-with-registers, the CLS embedding can be largely orthogonal to the convex combination of patch tokens, sharpening the CLS-vs-patch decoupling claim. \citet{marouani2026clspatch} proposed specialized processing paths for CLS and patch tokens to improve patch-feature quality for dense prediction. The dense-prediction literature \citep{ranftl2021dpt} has, for several years, used patch-token features rather than the CLS for spatially localized tasks, and the present paper does not contest that consensus. What this paper contributes, relative to that line of work, is a clinically-interpretable, large-scale ($n \approx 493$K aggregate), multi-cohort, multi-FM systematic localization of where on the spatial-extent axis the loss is overwhelming for chest radiography, plus a bbox-stratified clinical-task evaluation on a real CXR small-lesion cohort that converts the representational claim into a measurable size-stratified consequence.

\paragraph{Relation to prior CXR-specific work.} On chest radiography specifically, \citet{yang2025chexfound} introduced CheXFound, which makes the same diagnosis at the engineering level: relying on the CLS alone underutilizes the patch-token information learned during CXR pretraining, and a Global-and-Local Representations Integration (GLoRI) head that combines fine- and coarse-grained patch features with the CLS improves multilabel CXR classification over RAD-DINO, BiomedCLIP, and other frozen-CLS pipelines. \citet{li2025featurequality} reach a complementary empirical conclusion from a different angle: across eight medical and general-domain FMs benchmarked on chest X-ray, pre-trained embeddings are strong for global classification and segmentation of salient anatomy but ``perform poorly without significant fine-tuning'' for complex, subtle pathologies (e.g., pneumothorax segmentation), with subgroup analysis revealing that FMs rely on confounding shortcuts (such as chest tubes for pneumothorax) for classification --- a strategy that fails for precise localization. CheXFound demonstrates the engineering payoff, and \citet{li2025featurequality} document the empirical localization deficit on subtle pathology; the present paper provides the complementary mechanistic account at the embedding level (where exactly in the frozen forward pass the loss occurs, what representational properties drive it, and what spatial-extent threshold it operates at), shows that the recovery does not require a learned head when an ROI is supplied, and quantifies the size-stratified clinical-task gap on a separate small-lesion cohort. Where the medical-imaging-robustness literature has reached chest radiography \citep{cheng2025vlm}, it has measured task accuracy under artifact rather than the location of representational loss within the model. Recent embedding-probing methodology in general vision \citep{rashtchian2023substance} has not been applied to CXR at the scale and granularity needed to test the small-feature hypothesis. These are the specific gaps we address.

Concretely, we conduct controlled experiments on five frozen foundation models spanning three architecture families (ViT-B/14, ViT-7B/16, SigLIP2-So400m) and four pretraining paradigms (DINOv2, DINOv3, CLIP, SigLIP). Four families of small-scale perturbations: synthetic geometric patterns, localized isotropic Gaussian motion blur, physics-motivated directional motion blur (cranio-caudal and lateral, mimicking diaphragmatic and cardiac motion in portable acquisitions \citep{morin2018role}), and pathology-mimicking low-contrast patterns (narrow-band reticular, ground-glass), are injected into three large chest-radiography datasets (NIH-CXR14, MIMIC-CXR, Emory-CXR; aggregate $n = 492{,}724$ frontal radiographs). For every perturbation we compare three pooling modes derived from the same frozen forward pass (global CLS, patch-mean across all final-layer patch tokens, and patch-local restricted to tokens that intersect the perturbation), with raw-pixel and ImageNet-ResNet-50 baselines establishing pixel-level signal presence and a paired clean-vs-perturbed disease-classification protocol probing the downstream consequences. We then test whether the mechanism produces a measurable effect on real CXR small lesions, and whether a region-aware patch-pool pipeline can recover what the standard CLS path discards, on the ChestX-Det10 cohort \citep{liu2020chestxdet10}: 3{,}543 NIH-CXR14 images with 1{,}462 bbox annotations spanning three lesion-size classes (Calcification, Nodule, Mass; median bbox area ranging from $\approx 0.07\%$ to $\approx 2\%$ of total image area). The central empirical question is whether the embedding retains small-scale signal at all, where in the model it is lost, and whether the loss is recoverable in a region-aware downstream pipeline.

\section{Materials and Methods}
\label{sec:methods}

\subsection{Study Design}
\label{sec:design}

This retrospective study probes the small-scale-signal capacity of frozen ViT foundation-model embeddings in two complementary regimes. The first is the controlled-stimulus regime: small-scale perturbations injected into three large clean chest-radiography datasets function as standardized probes of known spatial extent and contrast. Linear probes on the resulting frozen embeddings, under three pooling modes (global CLS, patch-mean over all final-layer patch tokens, and patch-local restricted to tokens intersecting the perturbation), localize where in the representation small-scale signal is retained or discarded; a paired clean-vs-perturbed disease-classification protocol on cardiomegaly, edema, and lung-lesion provides the downstream check on globally-decided diseases. The second is the natural-stimulus regime: real small lesions on a real-CXR cohort with bbox annotations (ChestX-Det10), where the same three pooling modes are evaluated on lesion-classification tasks stratified by bbox area. The natural-stimulus regime tests whether the mechanism observed in the controlled-stimulus regime produces a measurable effect on real clinical-task classification, and whether region-aware patch pooling recovers it. All comparisons were pre-specified before analysis. Two ResNet-50 baselines are used: a fine-tuned ImageNet ResNet-50 as the pixel-level signal-verification baseline (Sec.~\ref{sec:resnet}, Sec.~\ref{sec:disc:resnet}), and a frozen DINO-pretrained ResNet-50 \citep{caron2021dino} as the matched architectural control isolating CNN vs ViT under shared SSL training and frozen inference (Sec.~\ref{sec:resnet}, Sec.~\ref{sec:res:isoblur}, Sec.~\ref{sec:disc:resnet}).

\subsection{Datasets}
\label{sec:datasets}

\subsubsection{NIH Chest X-ray 14 (NIH-CXR14)}
The publicly available NIH-CXR14 dataset \citep{wang2017chestxray8} comprising 112{,}120 frontal-view chest radiographs was used in its entirety to preserve real-world imaging diversity. Native resolution ($1{,}024\times 1{,}024$ px) was retained without resizing, padding, or intensity normalization to minimize interpolation artifacts and preserve anatomical geometry.

\subsubsection{MIMIC-CXR}
\label{sec:data:mimic}
The MIMIC-CXR-JPG dataset v2.0.0 \citep{johnson2019mimic} was downloaded from PhysioNet under credentialed access. After restriction to frontal views (ViewPosition $\in \{\text{AP, PA}\}$) and merging with the per-study CheXpert labels \citep{irvin2019chexpert}, 243{,}324 frontal radiographs remained. The official MIMIC-CXR-JPG train/validate/test split was used verbatim to ensure comparability with published work and to inherit the PhysioNet-enforced patient-level disjointness across splits. CheXpert uncertain labels ($-1$) were treated as negative following the U-Zero convention \citep{irvin2019chexpert}. To equalize statistical power across the three cohorts and reduce total compute by approximately one order of magnitude without loss of resolution for the reported between-model effect sizes, linear probes for MIMIC were trained on a pre-registered patient-disjoint stratified random subsample of the official training split ($n = 49{,}998$, seed = 42), with evaluation reported on the full official test split ($n = 11{,}072$ frontal). The exact manifest of included \texttt{dicom\_id}s, with \texttt{subject\_id}, \texttt{study\_id}, split assignment, and stratification key, is released as \texttt{mimic\_subsample\_ids.parquet} in the code repository. Images with heterogeneous dimensions were preprocessed via symmetric zero-padding followed by LANCZOS4 resampling to $1{,}024\times 1{,}024$ px.

\subsubsection{Emory-CXR}
The institutional Emory-CXR dataset was curated using defined inclusion criteria: anteroposterior (AP) view, no CLAHE applied, and balanced cardiomegaly representation. The dataset comprised 137{,}280 radiographs partitioned into training (80\%; $n=109{,}824$) and test (20\%; $n=27{,}456$) subsets. A separate 27{,}912-image subset was used for lung-lesion classification. Raw images with heterogeneous dimensions were preprocessed via symmetric zero-padding followed by LANCZOS4 resampling to $1{,}024\times 1{,}024$ px. Institutional IRB approval was obtained; all experiments on Emory-CXR ran on a PHI-compatible on-premises infrastructure and no Emory University images left that environment.

\paragraph{Spatial restriction of noise injection (all three datasets).} To ensure that small-perturbation probes test embedding behaviour on signal placed inside the anatomical region of interest rather than on signal placed in collimation bars, padded borders, or the corners of the radiograph, perturbations were placed only within the central 70\% of the image area on \emph{all three datasets} (NIH-CXR14, MIMIC-CXR, Emory-CXR). The restriction does not alter the noise-to-image area ratios reported in Sec.~\ref{sec:noise_ratios}, which are computed against the full image area in the standard way.

\subsubsection{ChestX-Det10 (Small-Lesion Bounding-Box Cohort)}
\label{sec:data:chestxdet10}
ChestX-Det10 \citep{liu2020chestxdet10} is a 3{,}543-image subset of NIH-CXR14 (3{,}001 train + 542 test) augmented in 2020 with instance-level bounding-box annotations across 10 thoracic-abnormality categories by three board-certified radiologists at Deepwise AI Lab. The bboxes were drawn de novo and are independent of the original NIH-CXR14 image-level pathology labels (which were extracted by NLP from radiology reports). We use the three lesion categories whose bbox-area distributions span the small-feature spatial-frequency band targeted by the synthetic-perturbation panel in Section~\ref{sec:perturbations}: \emph{Calcification} (347 bboxes / 201 positive images, median bbox $\approx 28\times 28$ px, $\approx 0.07\%$ of $1024 \times 1024$), \emph{Nodule} (955 bboxes / 384 positive images, median $\approx 45\times 45$ px, $\approx 0.19\%$), and \emph{Mass} (160 bboxes / 142 positive images, median $\approx 147\times 147$ px, $\approx 2.1\%$). Images were loaded directly from the Deepwise distribution at native $1{,}024 \times 1{,}024$ resolution. ChestX-Det10 inherits NIH-CXR14's image-side acquisition heterogeneity but is annotated independently of the synthetic-perturbation experiments above; no perturbations are applied in the ChestX-Det10 analysis (Section~\ref{sec:res:chestxdet10}), where the natural lesions themselves serve as the small-feature stimulus.

\subsection{Perturbation Design}
\label{sec:perturbations}

A controlled binary classification framework was used: images were assigned to a modified (``noisy'') or unmodified (``clean'') class with an enforced 50:50 class balance. Noise placement followed a deterministic pseudorandom procedure governed by SHA-256 hashing of the base seed concatenated with each image filename, ensuring reproducible yet non-repetitive spatial placement. Pixel intensities were adaptively sampled from the range of $\text{modal\_intensity} \pm 20$ gray levels, clipped to $[20, 235]$, preserving perceptual coherence with local image contrast. All perturbation injection was performed in an 8-bit intensity space: 8-bit source images (NIH-CXR14, MIMIC-CXR-JPG) were processed in their native bit depth, and 16-bit source images (Emory-CXR) were rescaled to 8-bit ($[0, 255]$) via linear contrast-stretch between the institution-provided window-level minimum and maximum before perturbation injection. This rescaling matches the JPEG-distributed bit depth of the two public cohorts and the input expected by each foundation model's HuggingFace processor (which normalises to a fixed mean/std on an 8-bit range), so the $[20, 235]$ clip range and $\pm 20$ gray-level sampling are comparable in perceptual contrast across all three datasets. The choice of a window-level linear contrast-stretch over a histogram-equalisation step was deliberate: \citet{dapamede2025dicom} show that for DICOM-sourced CXRs, histogram-equalisation enhancement produces models that overfit, lose external generalizability, and adopt shortcut features, whereas a VOI-LUT-style linear stretch preserves the radiometric relationships the encoder learned during pretraining. Bit-depth handling does not alter the noise-to-image area ratios reported below, which are spatial quantities.

\subsubsection{Synthetic Geometric Patterns}
Six geometric noise types were applied identically across all datasets: (a) circular noise (C1, C2): filled circles with radii of 1 and 2 px; (b) diagonal-line noise (L4, L8): 4$\times$4 and 8$\times$8 px tiles with diagonal strokes along the main or anti-diagonal; (c) square noise (S4, S8): 4$\times$4 and 8$\times$8 px patches with independently sampled pixel intensities, producing locally heterogeneous variation without directional structure.

\subsubsection{Localized Isotropic Gaussian Motion Blur}
\label{sec:isoblur}
A single isotropic Gaussian kernel of fixed size $21\times 21$ px ($\sigma$ auto-computed by OpenCV as $\sigma \approx 3.5$) was applied to the entire image to produce a fully blurred copy. A square patch of the blurred copy was then pasted back into the clean image at a pseudorandom position, so that the final image is globally clean except for one small blurred region. Two patch footprints were evaluated: 4$\times$4 px (16-pixel affected area) representing mild, localized motion blur, and 8$\times$8 px (64-pixel affected area) representing more substantial motion blur. The Gaussian kernel size ($21\times 21$) is held constant across both conditions; only the patch footprint that receives the blurred content varies.

\subsubsection{Physics-Motivated Directional Motion Blur}
\label{sec:dirblur}
Isotropic blur is a convenient proxy but does not capture the directional character of true acquisition motion. Respiratory excursion of the diaphragm produces a predominantly cranio-caudal point-spread function; cardiac wall motion produces a predominantly lateral point-spread function. We implemented linear motion-blur kernels parameterized by angle $\theta$ and kernel length $L$:
\begin{equation}
K_{\theta,L}(u,v) = \frac{1}{L}\cdot \mathbb{1}\!\left[(u,v) \in \mathrm{line}(\theta,L)\right],
\label{eq:linearkernel}
\end{equation}
where the line passes through the kernel center and the kernel is normalized to unit sum so that mean intensity is preserved. Two physically motivated angles were evaluated: $\theta = 90^\circ$ (cranio-caudal, mimicking diaphragmatic motion) and $\theta = 0^\circ$ (lateral, mimicking cardiac wall motion). Kernel lengths $L \in \{11,21,31\}$ px spanned exposure-equivalent extents; patch footprints $P \in \{16,32,64\}$ px varied the affected anatomical area. Each $(\theta, L, P)$ condition was evaluated on all three datasets for all five foundation models, yielding $2 \times 3 \times 3 \times 5 \times 3 = 270$ cells.

\subsubsection{Pathology-Mimicking Low-Contrast Patterns}
\label{sec:pathology}
To make the small-feature framing concrete on the diagnostic side, two simplified pathology-like perturbations whose spatial-frequency characteristics deliberately mirror real clinical findings \citep{henschke2006survival,hansell2008fleischner} were additionally evaluated: a narrow-band sinusoidal reticular grid of period 3 px (fine) or 6 px (coarse), amplitude-modulated to 8\% of the local patch mean, applied over $32\times 32$ or $64\times 64$ px patches (approximating fine interstitial reticulation); and a broad Gaussian intensity bump of $\sigma = 12$ px raising local mean intensity by 6\%, applied over a $64\times 64$ px patch (approximating ground-glass opacity). Amplitudes were chosen deliberately low to stay within plausible clinical contrast ranges.

\subsubsection{Noise-to-Image Area Ratios}
\label{sec:noise_ratios}
All tested perturbations are small relative to total image area ($1{,}024 \times 1{,}024 = 1{,}048{,}576$ px). Ratios are: C1 $\approx 0.0003\%$, C2 $\approx 0.0013\%$, S4/L4 $\approx 0.0015\%$, S8/L8 $\approx 0.006\%$, $4\times 4$ iso-blur $\approx 0.0015\%$, $8\times 8$ iso-blur $\approx 0.006\%$, $16 \times 16$ directional-blur $\approx 0.024\%$, $32\times 32 \approx 0.098\%$, $64 \times 64 \approx 0.39\%$, $32 \times 32$ reticular $\approx 0.098\%$, $64 \times 64$ ground-glass $\approx 0.39\%$. For reference, a 4 mm primary lung nodule on a 35 cm $\times$ 43 cm acquisition reconstructed to $1024 \times 1024$ occupies roughly 0.01\% of total image area, placing it within the lower half of the tested perturbation range.

\subsection{Foundation Models}
\label{sec:models}

Five frozen foundation models and two convolutional baselines were evaluated. Table~\ref{tab:models} summarizes architecture, pretraining domain, input resolution, and embedding dimensionality.

\begin{table*}[t]
\centering
\small
\begin{adjustbox}{max width=\textwidth}
\begin{tabular}{llllrrl}
\toprule
Model & HuggingFace ID & Architecture & Domain & Params & Emb.\ dim & Objective \\
\midrule
RAD-DINO      & \texttt{microsoft/rad-dino}                                       & ViT-B/14       & Radiology ($\sim$880K CXRs)  & 87M  & 768  & DINOv2-style \\
DINOv2-B/14   & \texttt{facebook/dinov2-base}                                     & ViT-B/14       & Natural ($\sim$142M)         & 87M  & 768  & DINOv2 \\
DINOv3 ViT-7B & \texttt{facebook/dinov3-vit7b16-pretrain-lvd1689m}                & ViT-7B/16      & Natural ($\sim$1.69B)        & 6.7B & 4{,}096 & DINOv3 \\
BiomedCLIP    & \texttt{microsoft/BiomedCLIP-PubMedBERT\_256-vit\_base\_patch16\_224} & ViT-B/16   & Biomedical ($\sim$15M)       & 86M  & 512  & CLIP \\
MedSigLIP     & \texttt{google/medsiglip-448}                                     & SigLIP2-So400m & Medical ($\sim$ billions)    & 428M & 1{,}152 & SigLIP \\
\midrule
ResNet-50 (FT)         & \texttt{ResNet50\_Weights.IMAGENET1K\_V2} & CNN & Natural (1.28M), fine-tuned & 25M & 2{,}048 & Supervised \\
DINO-ResNet-50 (frozen) & \texttt{dino\_resnet50\_pretrain.pth}    & CNN & Natural (1.28M), frozen SSL & 23M & 2{,}048 & DINO \citep{caron2021dino} \\
\bottomrule
\end{tabular}
\end{adjustbox}
\caption{\textbf{Foundation models evaluated.} DINOv2-B/14 serves as a size-matched control for RAD-DINO (same backbone, dimensionality, objective family --- only pretraining domain differs). The frozen DINO-ResNet-50 row is the matched architectural control (CNN backbone, same DINO SSL family, same frozen-inference + $L_2$-LR pipeline as the ViT CLS rows).}
\label{tab:models}
\end{table*}

All five foundation models were operated in frozen inference mode. The fine-tuned ImageNet ResNet-50 was fine-tuned end-to-end on each experimental dataset with the final fully connected layer replaced by a binary linear output; it serves as a pixel-level signal-verification baseline rather than a controlled architectural comparison. The DINO-pretrained ResNet-50 was held frozen at inference with the final classifier head replaced by an Identity layer, exposing the 2{,}048-d global-average-pooled feature, and run through the identical $L_2$-LR linear-probing pipeline as the ViT CLS rows (see Sec.~\ref{sec:resnet}).

\subsection{Feature Extraction and Classification}
\label{sec:features}

\subsubsection{Embedding Pooling Modes}
\label{sec:methods:pooling}
For each image, embeddings were extracted under three complementary pooling modes from the same forward pass:
\begin{description}[leftmargin=1.2em,itemsep=1pt]
  \item[CLS token.] The first output token of the final encoder layer, serving as the global image summary.\footnote{For consistency of nomenclature we use ``CLS'' throughout to denote the model's native global-aggregation output, even where the underlying mechanism is not strictly a classification token. RAD-DINO, DINOv2-B/14, DINOv3 ViT-7B, and BiomedCLIP use a prepended classification token in the canonical ViT manner. MedSigLIP (SigLIP2-So400m) uses a multi-head attention pooling (MAP) head instead of a CLS token; the ``CLS'' row for MedSigLIP refers to the MAP-pooled output of the vision encoder, which is the architectural analogue of the CLS token in that the same global-aggregation pathway is used to summarize all patch tokens into a single image-level vector. The mechanistic claim that small-scale signal is lost during this global aggregation step applies to either implementation; we use ``CLS-aggregation'' as shorthand for ``global-aggregation pooling'' across all five models.}
  \item[Patch-mean (mean pooling).] The arithmetic mean over all patch tokens of the final encoder layer, providing a global representation that does not privilege the CLS aggregation pathway.
  \item[Patch-local.] For perturbed images, the mean over only those patch tokens whose receptive field intersects the injected perturbation patch. Patch-local probing directly tests the patch-token-dilution hypothesis.
\end{description}
For both ResNet-50 baselines, the 2{,}048-dimensional global-average-pooled output was used (single pooling mode). For BiomedCLIP and MedSigLIP, the vision-encoder output (not the joint multimodal projection) was used.

\subsubsection{Linear Probing}
\label{sec:methods:probe}
All frozen embeddings were passed to $L_2$-regularized logistic-regression classifiers with class-balanced sample weights. Regularization strength $C$ was selected by 5-fold stratified cross-validation within the training set over the grid $C \in \{0.01, 0.1, 1.0, 10.0\}$; stratification was by the binary target label. The best-performing $C$ by mean validation AUC was refit on the full training set and evaluated once on the held-out test set. Solver was L-BFGS, maximum iterations 2{,}000, tolerance $10^{-4}$; features were $z$-scored via \texttt{StandardScaler} fit on training data only, with the same scaling applied to the test set. The primary random seed was fixed at 42 throughout.

\subsubsection{ResNet-50 Baselines}
\label{sec:resnet}
Two ResNet-50 baselines are used. The first, fine-tuned ImageNet ResNet-50, serves as a pixel-level signal-verification baseline. To maximize sensitivity for this verification purpose, all convolutional weights were fine-tuned end-to-end on each experimental dataset. Input size was $448 \times 448$ px with standard ImageNet normalization. This fine-tuned convolutional baseline differs from the frozen ViT foundation models in architecture, pretraining scale, domain, training objective, and inference mode, and is therefore \emph{not} used as the architectural test of the small-feature claim.

The second baseline, frozen DINO-pretrained ResNet-50 \citep{caron2021dino}, serves as the matched architectural control. It is a CNN backbone trained with the same DINO self-supervised objective family as DINOv2 and DINOv3, evaluated under the identical frozen inference + $L_2$-LR linear-probing pipeline as the ViT CLS embeddings, with the final classifier head replaced by an Identity layer to expose the 2{,}048-d global-average-pooled feature. Input size was $224 \times 224$ px with standard ImageNet normalization. By holding the training-objective family and the evaluation protocol fixed and varying only the architecture (CNN vs ViT) and the dimensionality, this baseline isolates the architectural contribution to the small-scale-signal-suppression finding.

\subsubsection{Raw-Pixel Baselines}
\label{sec:rawpixel}
The raw-pixel and the supervised ResNet-50 baselines (Sec.~\ref{sec:resnet}) jointly bracket the input-side discriminability envelope for each perturbation: they answer the question ``is the perturbation signal even present in the pixel data at this spatial scale?'' before we ask whether a frozen ViT embedding retains it. Three baselines are reported with deliberately different discriminative power. \emph{Raw-global} (downsampled-image linear probe) is a deliberately weak floor: it tests whether the perturbation produces a global pixel-statistic shift detectable by a linear classifier without any spatial prior. \emph{Raw-oracle} (location-known $32\times 32$ window, linear probe) is the linear pixel-level ceiling assuming perfect localization: it tests whether the perturbation is linearly separable from clean tissue inside its own footprint. \emph{ResNet-50 oracle} (fine-tuned non-linear CNN on the same window) is the supervised non-linear pixel-level ceiling: it tests whether the perturbation is detectable from pixel intensities at all, given a model expressive enough to learn the relevant non-linear features. The Raw-oracle / ResNet-50-oracle pair is informative precisely where the two diverge: raw-oracle at chance with ResNet-50-oracle far above it (e.g., iso-blur 8\,px on NIH-CXR14: raw-oracle 0.498 vs ResNet-50 0.983) means the signal is present in pixels but is non-linear in pixel space and thus not recoverable by a linear probe on raw pixels. This is exactly the situation in which a ViT CLS probe sitting at chance is uninformative about pixel-level signal presence, and the supervised ResNet-50 ceiling is the appropriate reference for ``is the signal there.''

Two raw-pixel logistic-regression baselines were evaluated. \emph{Raw-global}: the image was converted to grayscale and downsampled to $64\times 64$ via area interpolation (4{,}096 features). \emph{Raw-oracle}: a $32\times 32$ window centered on the known artifact patch location was extracted (1{,}024 features). Raw-oracle is a ceiling baseline assuming perfect location priors and is reported only to establish that the artifact is detectable from pixel intensities by a linear classifier; it is \emph{not} a model the foundation-model embedding could replicate, since the foundation models have no access to the patch coordinate at inference.

\subsubsection{Clean vs Perturbed Disease Classification}
\label{sec:cleanperturbed}
To assess whether embedding-level small-scale-signal loss translates into measurable downstream diagnostic effects, we trained each foundation model's disease classifier (cardiomegaly, edema, lung lesion) on clean training embeddings as standard. We then evaluated the same classifier under two test conditions: (a) the clean test set, and (b) the same test images after a single-patch perturbation injection (iso-blur 4 px and 8 px; directional blur, cranio-caudal and lateral at 32 and 64 px; reticular at 32 px fine; reticular at 64 px coarse; ground-glass at 64 px). Since both conditions share the same classifier and the same underlying test images, paired statistical tests apply directly. Cardiomegaly and edema are decided primarily on globally distributed evidence (cardiothoracic ratio, vascular redistribution) that does not occupy the spatial-frequency band these perturbations probe, so a near-zero $\Delta$AUC on those tasks is \emph{predicted} by the small-feature hypothesis and is not a refutation. The lesion-size-stratified analysis in Sec.~\ref{sec:res:chestxdet10} is the corresponding direct test, restricted to the bbox-annotated subset where ground-truth lesion area is known.

\subsubsection{ChestX-Det10 Probe Design (Natural-Lesion Regime)}
\label{sec:methods:cxd10}
For the natural-stimulus regime, two complementary probes were trained per (foundation model, lesion class) on ChestX-Det10 (Sec.~\ref{sec:data:chestxdet10}); no perturbations are injected.

\paragraph{Image-level binary detection.} An $L_2$-LR probe on the global CLS or patch-mean pool, with binary target ``image has class $X$ yes/no''. Trained on the official Deepwise train split (3{,}001 images), evaluated on the test split (542 images). For positive test images, test-set AUC was stratified by within-image median bbox area at the per-class median into small / large strata, with negatives held constant.

\paragraph{Bounding-box-level region-aware patch-pool comparison.} An $L_2$-LR probe on per-region features. Positive samples are real class-$X$ bboxes (one row per bbox); negative samples are matched-area random regions sampled from class-$X$-negative images at a 2:1 negatives-per-positive ratio. The matched-area, class-negative-image negatives are designed to (i) hold region scale constant between positives and negatives so the probe cannot exploit area as a shortcut, and (ii) keep the negative distribution disjoint from same-image residual lesion content; the 2:1 ratio gives the probe stable negative coverage at a moderate class imbalance without overwhelming the positive signal. The patch-local-vs-CLS gap reported in the corresponding results (Table~\ref{tab:chestxdet10_recovery}) is a within-design \emph{paired} contrast on identical (positives, negatives, train/test split) records, so the gap is invariant to changes in the negative-sampling ratio in expectation; what would shift under alternative negative pools (1:1, same-image non-bbox regions, or non-lesion-bearing anatomy) is the absolute AUC of each pool, not the patch-local-vs-CLS $\Delta$ that constitutes the headline claim. Three pool modes from the same frozen forward pass:
\begin{description}[leftmargin=1.4em,itemsep=1pt]
\item[CLS] image-level CLS pool (does not use the region; baseline).
\item[patch-mean (mean pooling)] image-level mean of patch tokens (does not use the region).
\item[patch-local] mean of final-layer patch tokens whose receptive field intersects the bbox (region-aware).
\end{description}
Train/test split inherits the official Deepwise split. Test-set AUC stratified by bbox area (per-class median split).

\paragraph{Per-cell paired inference.} Because CLS-pool and patch-local-pool predictions for each (FM, class) cell come from the same frozen forward pass on the same test bboxes, the patch-local-vs-CLS comparison is naturally paired. We carried out a per-cell paired bootstrap on $\Delta = \mathrm{AUC}(\text{patch-local}) - \mathrm{AUC}(\text{CLS})$ (10{,}000 resamples, identical bootstrap indices across pools so the within-pair correlation is preserved) and applied a Benjamini--Hochberg FDR correction across the matrix of 15 cells.

\subsection{Statistical Analysis}
\label{sec:stats}
Performance was reported using AUC and F1 at the best-F1 threshold; 95\% confidence intervals were computed via 1{,}000-iteration stratified bootstrap resampling on held-out test sets. For each paired model comparison on the same test set, we computed DeLong tests for correlated AUCs \citep{sun2014fast,delong1988comparing} and, as a distribution-free backup, label-swap permutation tests with 5{,}000 permutations. For clean-vs-perturbed disease classification, we computed paired-bootstrap 95\% CIs on $\Delta\text{AUC} = \text{AUC}_\text{clean} - \text{AUC}_\text{perturbed}$. Across the full comparison matrix, Benjamini--Hochberg FDR correction \citep{benjamini1995controlling} at $\alpha = 0.05$ was applied per model across all (dataset $\times$ disease $\times$ perturbation) cells (81 cells per model), with adjusted $p$-values reported alongside raw $p$-values.

\section{Results}
\label{sec:results}

\subsection{Disease Classification}
\label{sec:res:disease}

Table~\ref{tab:disease} reports disease-classification performance for all five foundation models on the three datasets. Across the 45 (model $\times$ disease $\times$ dataset) cells, AUC ranged from 0.642 (DINOv3 lung lesion on MIMIC) to 0.913 (MedSigLIP cardiomegaly on NIH), with mean 0.798. RAD-DINO and MedSigLIP achieved the highest mean AUCs (0.836 and 0.846 respectively across all 9 disease-dataset combinations); DINOv2 and DINOv3 had the lowest (0.768 and 0.773). All five models gave overlapping disease AUC ranges across datasets, satisfying the comparability condition necessary for the artifact-encoding analyses that follow: the dissociations reported in Sections~\ref{sec:res:isoblur}--\ref{sec:res:patchlevel} are therefore not driven by gross between-model differences in clinical-task capacity.

\begin{table}[pos=h]
\centering
\footnotesize
\begin{adjustbox}{max width=\linewidth}
\begin{tabular}{lllcccccc}
\toprule
           &             & & \multicolumn{2}{c}{NIH-CXR14} & \multicolumn{2}{c}{MIMIC-CXR} & \multicolumn{2}{c}{Emory-CXR} \\
 \cmidrule(lr){4-5} \cmidrule(lr){6-7} \cmidrule(lr){8-9}
Task       & Model       & & AUC & F1 & AUC & F1 & AUC & F1 \\
\midrule
\multirow{5}{*}{Cardiomegaly}
 & RAD-DINO   && 0.900 & 0.342 & 0.797 & 0.506 & 0.826 & 0.769 \\
 & DINOv2-B   && 0.818 & 0.198 & 0.752 & 0.463 & 0.755 & 0.726 \\
 & DINOv3     && 0.836 & 0.253 & 0.763 & 0.477 & 0.805 & 0.756 \\
 & BiomedCLIP && 0.877 & 0.294 & 0.776 & 0.488 & 0.802 & 0.751 \\
 & MedSigLIP  && 0.913 & 0.359 & 0.812 & 0.527 & 0.855 & 0.791 \\
\midrule
\multirow{5}{*}{Edema}
 & RAD-DINO   && 0.899 & 0.270 & 0.876 & 0.520 & 0.825 & 0.545 \\
 & DINOv2-B   && 0.875 & 0.231 & 0.853 & 0.480 & 0.775 & 0.485 \\
 & DINOv3     && 0.858 & 0.213 & 0.851 & 0.481 & 0.797 & 0.508 \\
 & BiomedCLIP && 0.883 & 0.239 & 0.858 & 0.490 & 0.795 & 0.507 \\
 & MedSigLIP  && 0.911 & 0.267 & 0.889 & 0.549 & 0.843 & 0.570 \\
\midrule
\multirow{5}{*}{Lung Lesion}
 & RAD-DINO   && 0.782 & 0.320 & 0.758 & 0.214 & 0.806 & 0.242 \\
 & DINOv2-B   && 0.703 & 0.225 & 0.652 & 0.094 & 0.720 & 0.112 \\
 & DINOv3     && 0.706 & 0.233 & 0.642 & 0.124 & 0.698 & 0.097 \\
 & BiomedCLIP && 0.690 & 0.215 & 0.686 & 0.189 & 0.738 & 0.141 \\
 & MedSigLIP  && 0.771 & 0.297 & 0.751 & 0.205 & 0.798 & 0.210 \\
\bottomrule
\end{tabular}
\end{adjustbox}
\caption{\textbf{Foundation-model disease-classification performance.} Point estimates of AUC and F1 at best-F1 threshold; 95\% bootstrap confidence intervals are reported in Supplementary Table~S1. All five foundation models produce overlapping disease AUC across datasets (range 0.642--0.913, mean 0.798).}
\label{tab:disease}
\end{table}

\subsection{Synthetic Geometric Noise Detection}
\label{sec:res:geometric}

The smallest perturbations in the panel --- single-pixel circles, 4-pixel and 8-pixel diagonal lines and squares --- sit at the lower boundary of the spatial-frequency band of interest (0.0003\%--0.006\% of image area). Across all 90 (model $\times$ pattern $\times$ dataset) cells (5 models $\times$ 6 patterns $\times$ 3 datasets), CLS-probe AUC ranged from 0.500 (chance) to 0.604 (MedSigLIP S8 on NIH); 76 of 90 cells had AUC $\leq 0.51$. Table~\ref{tab:geometric} reports a representative summary; the full matrix is in Supplementary Table~S2.

\begin{table*}[pos=ht]
\centering
\footnotesize
\begin{adjustbox}{max width=\textwidth}
\begin{tabular}{lcccccc}
\toprule
& \multicolumn{6}{c}{NIH-CXR14 AUC (CLS pooling)} \\
\cmidrule(lr){2-7}
Model      & C1 & C2 & L4 & L8 & S4 & S8 \\
\midrule
RAD-DINO   & 0.500 & 0.505 & 0.500 & 0.500 & 0.500 & 0.503 \\
DINOv2-B   & 0.500 & 0.517 & 0.500 & 0.501 & 0.501 & 0.507 \\
DINOv3     & 0.501 & 0.518 & 0.500 & 0.501 & 0.502 & 0.509 \\
BiomedCLIP & 0.500 & 0.500 & 0.500 & 0.500 & 0.500 & 0.500 \\
MedSigLIP  & 0.524 & 0.567 & 0.506 & 0.511 & 0.526 & 0.604 \\
\midrule
ResNet-50 (fine-tuned, pixel-signal verification)
           & 0.978 & 0.989 & 0.978 & 0.985 & 0.972 & 0.992 \\
\bottomrule
\end{tabular}
\end{adjustbox}
\caption{\textbf{Synthetic geometric noise detection on NIH-CXR14.} Linear-probe AUC on frozen CLS embeddings (foundation models) or fine-tuned global-average-pooled features (ResNet-50). The corresponding MIMIC and Emory-CXR matrices follow the same pattern (Supplementary Table~S2): all five foundation models give near-random CLS detection across the 90-cell panel, while ResNet-50 achieves AUC 0.93--0.99, confirming pixel-level discriminability is satisfied at the relevant spatial scale and locating the loss to the embedding rather than to the input.}
\label{tab:geometric}
\end{table*}

\subsection{Localized Isotropic Motion Blur Detection}
\label{sec:res:isoblur}

Table~\ref{tab:isoblur} presents detection performance for localized isotropic Gaussian motion blur on all three datasets. The $21 \times 21$ Gaussian kernel ($\sigma \approx 3.5$) was held constant; only the patch footprint varied (4$\times$4 and 8$\times$8 px; see Sec.~\ref{sec:isoblur}). The result is a clean small-feature null across all five foundation models on all three datasets: every cell falls in the range AUC 0.500--0.503, with 95\% bootstrap CIs spanning chance for every cell. The frozen DINO-pretrained ResNet-50 architectural control reproduces the same chance floor on both NIH-CXR14 (AUC 0.500 [0.488, 0.513] at $P=4$, 0.500 [0.488, 0.514] at $P=8$) and Emory-CXR (AUC 0.500 [0.487, 0.512] and 0.500 [0.488, 0.513]), establishing that the chance-floor behaviour is not exclusive to the ViT architecture and is reproduced by a CNN trained under the same DINO self-supervised objective and evaluated under the same frozen $L_2$-LR pipeline. We note that this control is single-objective-family (DINO-style SSL only); replication on frozen MoCo-v3, BYOL, or MAE CNN backbones would be needed before the loss can be attributed to frozen-SSL global pooling in general rather than to the DINO objective specifically (Sec.~\ref{sec:limitations}). Raw-pixel oracle (32$\times$32 window centered on the known patch location) is itself at chance for these conditions on all three datasets (NIH 0.498 / 0.498; MIMIC 0.510 / 0.505; Emory-CXR 0.497 / 0.491), confirming that iso-blur 4--8 px sits at the lower edge of the input-side discriminability envelope. ResNet-50 oracle reaches AUC 0.685 / 0.983 on NIH at $P=4 / P=8$, providing the supervised pixel-positive ceiling that confirms signal presence at the larger of the two footprints.

\begin{table*}[pos=ht]
\centering
\footnotesize
\begin{adjustbox}{max width=\textwidth}
\begin{tabular}{lcccccc}
\toprule
            & \multicolumn{2}{c}{NIH-CXR14} & \multicolumn{2}{c}{MIMIC-CXR} & \multicolumn{2}{c}{Emory-CXR} \\
 \cmidrule(lr){2-3} \cmidrule(lr){4-5} \cmidrule(lr){6-7}
Model / probe   & 4$\times$4 & 8$\times$8 & 4$\times$4 & 8$\times$8 & 4$\times$4 & 8$\times$8 \\
\midrule
RAD-DINO (CLS)         & 0.500 & 0.500 & 0.500 & 0.500 & 0.500 & 0.500 \\
DINOv2-B (CLS)         & 0.500 & 0.500 & 0.501 & 0.502 & 0.500 & 0.500 \\
DINOv3 (CLS)           & 0.500 & 0.500 & 0.501 & 0.503 & 0.500 & 0.501 \\
BiomedCLIP (CLS)       & 0.500 & 0.500 & 0.500 & 0.500 & 0.500 & 0.500 \\
MedSigLIP (CLS)        & 0.500 & 0.500 & 0.501 & 0.502 & 0.500 & 0.500 \\
\midrule
DINO-ResNet-50 (frozen, architectural control) $\dagger$
                       & 0.500 & 0.500 & --- & --- & 0.500 & 0.500 \\
Raw-pixel global       & 0.500 & 0.500 & 0.500 & 0.500 & 0.500 & 0.500 \\
Raw-pixel oracle (32$\times$32 around patch) & 0.498 & 0.498 & 0.510 & 0.505 & 0.497 & 0.491 \\
ResNet-50 oracle (fine-tuned, pixel pos.\ ctrl) $\ddagger$ & 0.685 & 0.983 & --- & --- & --- & --- \\
\bottomrule
\end{tabular}
\end{adjustbox}
\caption{\textbf{Localized isotropic motion blur, all five foundation models on three datasets.} Linear-probe AUC on frozen embeddings; Gaussian kernel size 21$\times$21 px held constant, only the patch footprint varies. All 30 (model $\times$ patch $\times$ dataset) ViT cells coincide at chance. $\dagger$ DINO-ResNet-50 \citep{caron2021dino} architectural control; identical $L_2$-LR pipeline; reproduces the chance floor on both NIH-CXR14 and Emory-CXR. $\ddagger$ Supervised pixel positive control. Full breakdown with 95\% CIs in Supplementary Table~S4.}
\label{tab:isoblur}
\end{table*}

\subsection{Directional (Physics-Motivated) Motion Blur Detection}
\label{sec:res:dirblur}

Table~\ref{tab:dirblur} reports the representative directional-blur condition ($L=21$, $P=64$, both directions) for all five foundation models on all three datasets. The full $2 \times 3 \times 3 = 18$-cell sweep per (model, dataset) pair (270 cells total) is in Supplementary Table~S3. At the smallest footprint ($P=16$ px), all five ViTs are at chance (AUC 0.500--0.506 across the 30 corresponding cells); detection rises monotonically with $P$, and a clear ordering emerges only at $P=64$: BiomedCLIP and RAD-DINO are held in the 0.52--0.55 band, DINOv2-B/14 reaches 0.60--0.62, and DINOv3 (4{,}096-d, ViT-7B) and MedSigLIP (1{,}152-d, SigLIP2-So400m) reach 0.69--0.77 on cranio-caudal blur. The frozen DINO-ResNet-50 architectural control reaches AUC 0.600 on cranio-caudal $P=64$ on Emory-CXR, comparable to DINOv2-B/14 and substantially below DINOv3/MedSigLIP, supporting the interpretation that capacity-driven encoding rises above the chance floor only above a spatial-extent threshold. Critically, even DINOv3's recovery (max AUC $\approx 0.77$ at $L=31$, $P=64$) does not match the patch-local oracle's AUC $\approx 1.0$ on the same perturbations (Sec.~\ref{sec:res:patchlevel}); the signal is in patch tokens but only partially recovered by global aggregation even at the upper edge.

\begin{table*}[pos=ht]
\centering
\footnotesize
\begin{adjustbox}{max width=\textwidth}
\begin{tabular}{lcccccc}
\toprule
& \multicolumn{2}{c}{NIH-CXR14 ($L=21$, $P=64$)} & \multicolumn{2}{c}{MIMIC-CXR ($L=21$, $P=64$)} & \multicolumn{2}{c}{Emory-CXR ($L=21$, $P=64$)} \\
\cmidrule(lr){2-3} \cmidrule(lr){4-5} \cmidrule(lr){6-7}
Model      & cranio & lateral & cranio & lateral & cranio & lateral \\
\midrule
RAD-DINO   & 0.537 & 0.541 & 0.541 & 0.545 & 0.545 & 0.549 \\
DINOv2-B   & 0.611 & 0.602 & 0.623 & 0.613 & 0.619 & 0.607 \\
DINOv3     & 0.726 & 0.714 & 0.732 & 0.720 & 0.717 & 0.700 \\
BiomedCLIP & 0.527 & 0.526 & 0.528 & 0.526 & 0.526 & 0.525 \\
MedSigLIP  & 0.688 & 0.697 & 0.770 & 0.749 & 0.710 & 0.722 \\
\bottomrule
\end{tabular}
\end{adjustbox}
\caption{\textbf{Directional motion-blur detection at $L = 21$, $P = 64$ (representative condition).} Linear-probe AUC on frozen CLS embeddings; cranio-caudal ($\theta = 90^\circ$) mimics diaphragmatic motion, lateral ($\theta = 0^\circ$) mimics cardiac motion. The 0.500--0.566 vs 0.700--0.788 spread between low-capacity (RAD-DINO, BiomedCLIP) and high-capacity (DINOv3, MedSigLIP) ViTs at the largest footprint isolates the dimensionality contribution to encoding capacity above the spatial-extent threshold. Full $2 \times 3 \times 3$ sweep for all 5 models on all 3 datasets in Supplementary Table~S3.}
\label{tab:dirblur}
\end{table*}

\subsection{Pathology-Mimicking Low-Contrast Patterns}
\label{sec:res:pathology}

Table~\ref{tab:pathology} reports detection performance for the two pathology-mimicking perturbations (reticular fine $P=32$, ground-glass $P=64$; Sec.~\ref{sec:pathology}). All five foundation models detected the perturbations at AUC 0.500--0.588 on NIH-CXR14, 0.502--0.564 on MIMIC-CXR, and 0.500--0.571 on Emory-CXR, close to the 0.5 chance floor across all 30 (model $\times$ pattern $\times$ dataset) cells. The raw-pixel oracle and ResNet-50 oracle baselines confirm pixel-level discriminability is satisfied for these conditions (raw-oracle reticular fine 32 AUC 1.000 on all three datasets; raw-oracle ground-glass 64 AUC 0.741 / 0.676 / 0.744 on NIH / MIMIC / Emory-CXR), localizing the loss to the embedding rather than the input.

\begin{table}[pos=h]
\centering
\footnotesize
\begin{adjustbox}{max width=\linewidth}
\begin{tabular}{lcccccc}
\toprule
& \multicolumn{2}{c}{NIH-CXR14} & \multicolumn{2}{c}{MIMIC-CXR} & \multicolumn{2}{c}{Emory-CXR} \\
\cmidrule(lr){2-3} \cmidrule(lr){4-5} \cmidrule(lr){6-7}
Model      & Ret.\ fine 32 & GG 64 & Ret.\ fine 32 & GG 64 & Ret.\ fine 32 & GG 64 \\
\midrule
RAD-DINO   & 0.509 & 0.588 & 0.504 & 0.540 & 0.505 & 0.557 \\
DINOv2-B   & 0.502 & 0.551 & 0.507 & 0.538 & 0.501 & 0.548 \\
DINOv3     & 0.512 & 0.576 & 0.514 & 0.560 & 0.510 & 0.570 \\
BiomedCLIP & 0.500 & 0.519 & 0.502 & 0.511 & 0.500 & 0.518 \\
MedSigLIP  & 0.527 & 0.572 & 0.510 & 0.542 & 0.512 & 0.560 \\
\bottomrule
\end{tabular}
\end{adjustbox}
\caption{\textbf{Pathology-mimicking pattern detection on all three datasets.} Linear-probe AUC on frozen CLS embeddings. Reticular fine = period 3 px, amplitude 8\% of local mean; ground-glass = Gaussian bump $\sigma = 12$ px, amplitude 6\%.}
\label{tab:pathology}
\end{table}

\subsection{Patch-Level Probing: CLS vs Patch-Mean vs Patch-Local}
\label{sec:res:patchlevel}

This is the centerpiece result. Across all 90 (model $\times$ perturbation $\times$ dataset) cells (5 models $\times$ 6 perturbations $\times$ 3 datasets), patch-local probes recovered AUC $\geq 0.99$ on every cell where the CLS probe was at or near chance. Per-model mean gaps $\Delta = \text{AUC}(\text{patch-local}) - \text{AUC}(\text{CLS})$, computed across 18 (perturbation $\times$ dataset) cells per model, were:

\begin{itemize}[leftmargin=1.6em,itemsep=1pt]
\item RAD-DINO: $+0.474$ (range $+0.412$ to $+0.500$)
\item DINOv2-B/14: $+0.453$ (range $+0.376$ to $+0.499$)
\item DINOv3: $+0.413$ (range $+0.268$ to $+0.500$)
\item BiomedCLIP: $+0.488$ (range $+0.472$ to $+0.500$)
\item MedSigLIP: $+0.412$ (range $+0.230$ to $+0.497$)
\end{itemize}

The signal is fully present in the final-layer patch tokens; the CLS aggregation step is where it is lost. Table~\ref{tab:pooling} gives a representative slice on NIH-CXR14; the full per-dataset breakdown is in Supplementary Table~S5.

\begin{table*}[pos=h]
\centering
\footnotesize
\begin{adjustbox}{max width=\textwidth}
\begin{tabular}{llccccc}
\toprule
Perturbation & Model & CLS AUC & Patch-mean AUC & Patch-local AUC & $\Delta$ (local $-$ CLS) \\
\midrule
\multirow{5}{*}{iso-blur 4 px}
 & RAD-DINO   & 0.500 & 0.500 & 1.000 & $+0.500$ \\
 & DINOv2-B   & 0.500 & 0.500 & 0.999 & $+0.499$ \\
 & DINOv3     & 0.500 & 0.500 & 1.000 & $+0.500$ \\
 & BiomedCLIP & 0.500 & 0.500 & 0.998 & $+0.498$ \\
 & MedSigLIP  & 0.500 & 0.500 & 0.994 & $+0.494$ \\
\midrule
\multirow{5}{*}{dir-cranio 64 px}
 & RAD-DINO   & 0.537 & 0.595 & 1.000 & $+0.463$ \\
 & DINOv2-B   & 0.611 & 0.620 & 1.000 & $+0.388$ \\
 & DINOv3     & 0.726 & 0.726 & 1.000 & $+0.274$ \\
 & BiomedCLIP & 0.527 & 0.528 & 1.000 & $+0.473$ \\
 & MedSigLIP  & 0.688 & 0.713 & 1.000 & $+0.312$ \\
\midrule
\multirow{5}{*}{reticular fine 32 px}
 & RAD-DINO   & 0.509 & 0.525 & 1.000 & $+0.491$ \\
 & DINOv2-B   & 0.502 & 0.502 & 1.000 & $+0.498$ \\
 & DINOv3     & 0.512 & 0.509 & 1.000 & $+0.488$ \\
 & BiomedCLIP & 0.500 & 0.500 & 0.999 & $+0.499$ \\
 & MedSigLIP  & 0.527 & 0.533 & 0.999 & $+0.472$ \\
\bottomrule
\end{tabular}
\end{adjustbox}
\caption{\textbf{Three-pooling comparison on NIH-CXR14 (representative slice).} ``CLS'' = final-layer CLS token; ``patch-mean'' = mean over all final-layer patch tokens; ``patch-local'' = mean over final-layer patch tokens whose receptive field intersects the injected perturbation. All three pools come from the same frozen forward pass. Across all 90 (model $\times$ perturbation $\times$ dataset) cells, patch-local recovers AUC $\geq 0.99$ on every cell with a chance-level CLS probe. Per-dataset breakdown for MIMIC-CXR and Emory-CXR in Supplementary Table~S5.}
\label{tab:pooling}
\end{table*}

\begin{figure*}[pos=ht]
\centering
\includegraphics[width=\textwidth]{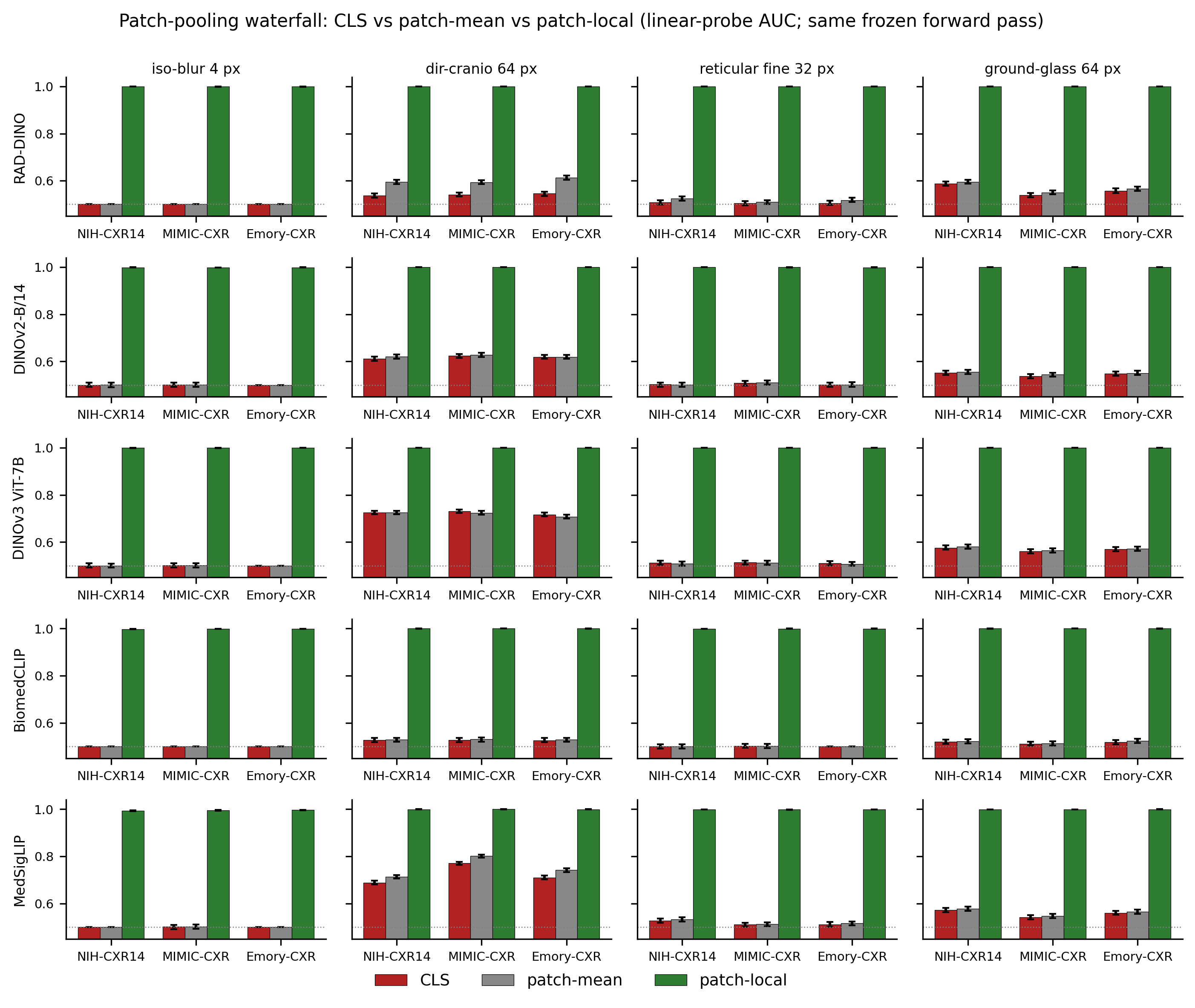}
\caption{\textbf{Patch-pooling waterfall for representative perturbations.} Per (model, perturbation, dataset) cell, three bars: CLS, patch-mean, patch-local. The patch-local bar reaches AUC $\geq 0.99$ on every cell with a chance-level CLS bar, directly localizing the loss of small-scale signal to the global-aggregation step of the frozen forward pass. The patch-mean bar is indistinguishable from CLS at the chance floor for every iso-blur and reticular-fine cell, confirming that the loss is a property of any whole-image aggregation rather than of the CLS pathway specifically.}
\label{fig:pooling_waterfall}
\end{figure*}

\begin{figure*}[pos=ht]
\centering
\includegraphics[width=\textwidth]{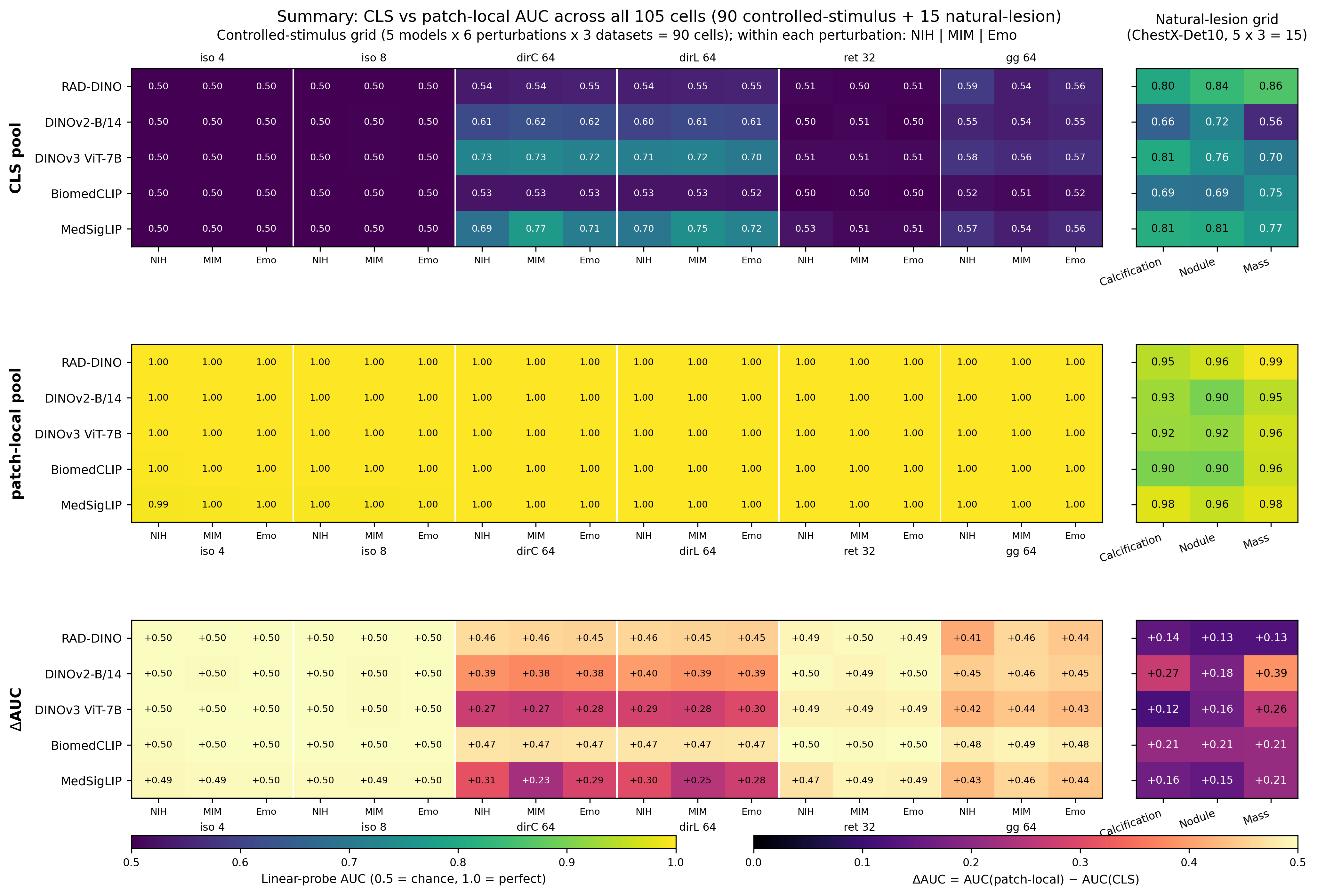}
\caption{\textbf{Summary: CLS vs patch-local AUC across all experimental conditions.} Heat-map matrix of linear-probe AUC across the 90 controlled-stimulus cells (5 foundation models $\times$ 6 representative perturbations $\times$ 3 datasets) and the 15 natural-lesion cells (5 FMs $\times$ 3 ChestX-Det10 classes). Top row: CLS-pool AUC clusters around 0.50 across the full controlled-stimulus grid; the only cells lifting off the chance floor are directional-blur conditions at $P \geq 32$ for the higher-capacity DINOv3 and MedSigLIP. Bottom row: patch-local AUC clusters near 1.0 on every controlled-stimulus cell and at $0.899$--$0.991$ on every ChestX-Det10 cell. The within-pair difference (right panel) ranges from $+0.119$ to $+0.500$ across the matrix, with a uniform visual signature: the loss is consistent across architectures, pretraining domains, and objectives, and the patch-local recovery is consistent across the controlled-stimulus and natural-lesion regimes. Cells are coloured on a 0.5--1.0 AUC scale; full per-cell numerics in Tables~\ref{tab:isoblur}, \ref{tab:dirblur}, \ref{tab:pathology}, \ref{tab:pooling}, \ref{tab:chestxdet10_recovery} and Supplementary Tables~S2--S5, S10--S11.}
\label{fig:summary_grid}
\end{figure*}

\subsection{Raw-Pixel Baselines}
\label{sec:res:rawpixel}

Table~\ref{tab:rawpixel} shows raw-pixel logistic-regression baselines bracketing the input-side discriminability floor. The raw-oracle baseline ($32 \times 32$ window at the known perturbation location) reaches AUC 1.000 on the reticular pattern across all three datasets and AUC 0.676--0.744 on ground-glass, confirming pixel-level discriminability for the spatially-broader perturbations. On the smallest perturbations (iso-blur 4--8 px), even the oracle window is at chance, identifying the lower edge of the discriminability envelope. The raw-global baseline ($64 \times 64$ downsampled full image) is at chance throughout, showing that whatever signal is present is spatially local. Together with the ResNet-50 baselines in Sec.~\ref{sec:res:geometric} and \ref{sec:res:isoblur}, this rules out task difficulty as an explanation for the foundation-model results: where the input contains a discriminable signal at the relevant spatial scale, the absence of that signal in the foundation-model CLS embedding is a property of the representation, not of the input.

\begin{table*}[pos=h]
\centering
\footnotesize
\begin{adjustbox}{max width=\textwidth}
\begin{tabular}{lcccccc}
\toprule
            & \multicolumn{2}{c}{NIH-CXR14} & \multicolumn{2}{c}{MIMIC-CXR} & \multicolumn{2}{c}{Emory-CXR} \\
\cmidrule(lr){2-3} \cmidrule(lr){4-5} \cmidrule(lr){6-7}
Perturbation & Raw-global & Raw-oracle & Raw-global & Raw-oracle & Raw-global & Raw-oracle \\
\midrule
iso-blur 4 px               & 0.500 & 0.498 & 0.500 & 0.510 & 0.500 & 0.497 \\
iso-blur 8 px               & 0.500 & 0.498 & 0.500 & 0.505 & 0.500 & 0.491 \\
dir-cranio 64 px            & 0.500 & 0.507 & 0.500 & 0.498 & 0.500 & 0.495 \\
reticular fine 32 px        & 0.500 & 1.000 & 0.500 & 0.999 & 0.500 & 1.000 \\
ground-glass 64 px          & 0.501 & 0.741 & 0.501 & 0.676 & 0.502 & 0.744 \\
\bottomrule
\end{tabular}
\end{adjustbox}
\caption{\textbf{Raw-pixel logistic-regression baselines, all three datasets.} Raw-global = $64 \times 64$ downsampled full image (4{,}096 features). Raw-oracle = $32 \times 32$ window at the known perturbation location (1{,}024 features). 95\% CIs in Supplementary Table~S6.}
\label{tab:rawpixel}
\end{table*}

\subsection{Downstream Disease Classification Under Perturbation}
\label{sec:res:clinicalloop}

Table~\ref{tab:clinicalloop} summarizes the paired clean-vs-perturbed disease-classification protocol across all (dataset $\times$ disease $\times$ perturbation) cells, BH-FDR adjusted per model at $\alpha = 0.05$ (81 cells per model = 3 datasets $\times$ 3 diseases $\times$ 9 perturbations). Mean $\Delta$AUC per model was at most $+0.00126$; counts of cells significant at BH-FDR 0.05 ranged from 0/81 (RAD-DINO, BiomedCLIP) to 7/81 (MedSigLIP). Under the small-feature hypothesis, this near-zero $\Delta$AUC pattern is the \emph{predicted} outcome on these specific disease tasks: cardiomegaly is decided by cardiothoracic ratio; edema by diffuse vascular redistribution; the NIH-CXR14 \texttt{lung\_lesion} label aggregates \texttt{Mass} and \texttt{Nodule} across the full size range. None of these diagnostic signals lies in the small-scale, low-contrast band the perturbations probe. The corresponding direct test on a small-feature-bearing clinical task is reported in Sec.~\ref{sec:res:chestxdet10}.

\begin{table*}[pos=h]
\centering
\footnotesize
\begin{adjustbox}{max width=\textwidth}
\begin{tabular}{lcccc}
\toprule
Model       & Mean $\Delta$AUC [95\% CI] & Cells sig.\ at BH-FDR 0.05 & Total cells & \% sig. \\
\midrule
RAD-DINO    & $+0.00024\,[-0.00010, +0.00058]$ & 0  & 81 & 0.0\% \\
DINOv2-B/14 & $+0.00026\,[-0.00012, +0.00064]$ & 3  & 81 & 3.7\% \\
DINOv3      & $+0.00021\,[-0.00018, +0.00060]$ & 2  & 81 & 2.5\% \\
BiomedCLIP  & $-0.00003\,[-0.00031, +0.00026]$ & 0  & 81 & 0.0\% \\
MedSigLIP   & $+0.00126\,[+0.00064, +0.00188]$ & 7  & 81 & 8.6\% \\
\bottomrule
\end{tabular}
\end{adjustbox}
\caption{\textbf{Clean vs perturbed disease classification: aggregate per-model summary.} 81 cells = 3 datasets $\times$ 3 diseases $\times$ 9 perturbations. $\Delta$AUC = AUC$_\mathrm{clean}$ $-$ AUC$_\mathrm{perturbed}$. Significance count is at BH-FDR 0.05 per model across the 81-cell matrix. Full $\sim$405-row table in Supplementary Table~S8.}
\label{tab:clinicalloop}
\end{table*}

\subsection{Embedding-Space Separation (UMAP, Silhouette)}
\label{sec:res:umap}

UMAP \citep{mcinnes2018umap} projections of clean vs perturbed CLS embeddings for each (model, perturbation, dataset) tuple are shown in Supplementary Figure~S2. Quantitative separation measures \citep{rousseeuw1987silhouettes} on the same pairs are summarized in Supplementary Table~S7. The cells with the largest pooled-covariance Mahalanobis distances between clean and perturbed centroids (DINOv3 / MedSigLIP at dir-cranio $P=64$, $\approx 0.7$--0.8) coincide with the cells where those models' CLS probes also achieve the highest detection AUC, supporting cross-method convergence between linear-probe AUC and embedding-space geometry.

\subsection{ChestX-Det10 Small-Lesion Bounding-Box-Stratified Analysis}
\label{sec:res:chestxdet10}

The synthetic-perturbation results in Sections~\ref{sec:res:geometric}--\ref{sec:res:patchlevel} establish the mechanistic claim on a controlled stimulus: across five frozen ViT foundation models and three large chest-radiography datasets, CLS-pool embeddings are at chance for sub-1\%-area perturbations, while patch-local pools on the same forward pass recover near-perfect linear-probe AUC. We turn now to the corresponding clinical-task test on a real-CXR small-lesion cohort. The two questions are: (i) does the mechanism produce a measurable size-stratified deficit in real-lesion classification at the CLS pool, and (ii) does the same patch-local recovery extend from controlled perturbation patches to natural lesion bboxes. Probe design (positives, matched-area negatives, three pool modes, paired-bootstrap inference) is described in Methods, Sec.~\ref{sec:methods:cxd10}.

\subsubsection{Image-Level Detection: A CLS-Pool Size-Stratified Gap}
\label{sec:cxd10:imagelevel}

Across all 5 foundation models $\times$ 3 lesion classes (15 cells), the size-stratified test on CLS-pool embeddings reveals a positive within-class small-vs-large stratum gap on 14/15 cells (Supplementary Table~S10); the only cell with a non-positive gap is DINOv2-B/14 Mass ($-0.020$), which sits in the lowest-power class stratum (Mass class has only 31 test positives, splitting to 15--16 per stratum at the median; the minimum detectable stratum-AUC difference at $\alpha = 0.05$ is $\approx 0.33$ at this $n$, well above the $|\Delta|= 0.020$ observed). Across the 14 cells with a positive gap, mean image-level CLS stratum gap (large $-$ small AUC) is $+0.118$; the gap reaches up to $+0.243$ (RAD-DINO, Nodule: small $=0.611$ vs large $=0.854$) and is consistently large for the most challenging cells (e.g., MedSigLIP Nodule small $=0.627$ vs large $=0.844$, gap $+0.218$; MedSigLIP Calcification small $=0.631$ vs large $=0.831$, gap $+0.200$). For Calcification, every model's small-stratum CLS AUC sits between 0.574 and 0.689 and every model's large-stratum AUC sits between 0.686 and 0.831; for Nodule, small 0.605--0.661 vs large 0.689--0.854; for Mass, small 0.627--0.783 vs large 0.665--0.936. The dilution mechanism observed in the controlled-stimulus regime thus produces a measurable size-stratified disadvantage on natural lesions in the natural-stimulus regime; the bbox-level patch-local probe in Sec.~\ref{sec:cxd10:bboxlevel}, decided on the full positive count rather than the median split, is the corresponding test that is not subject to stratum power loss.

\subsubsection{Bounding-Box-Level Region-Aware Pooling: Recovery}
\label{sec:cxd10:bboxlevel}

This is the centerpiece result of the section. Across all 15 (FM $\times$ class) cells, patch-local AUC exceeds CLS-pool AUC on the same forward pass by $+0.119$ to $+0.387$ (Table~\ref{tab:chestxdet10_recovery}). Per-FM mean patch-local-vs-CLS gap, averaged over the three classes:
\begin{itemize}[leftmargin=1.6em,itemsep=1pt]
\item DINOv2-B/14: $+0.277$ (largest; natural-image SSL benefits most from region-awareness)
\item BiomedCLIP: $+0.212$
\item DINOv3 ViT-7B: $+0.182$
\item MedSigLIP: $+0.175$
\item RAD-DINO: $+0.134$ (smallest; radiology SSL already encodes some lesion-region features globally)
\end{itemize}
Patch-local pool reaches AUC $\geq 0.899$ on every cell and AUC $\geq 0.94$ on Mass for every model (minima: BiomedCLIP Nodule $0.899$ across all cells; DINOv2 Mass $0.948$ within the Mass class). The lesion content is fully recoverable from the patch-token representation with bbox-aware ROI pooling, on the same forward pass that gave a size-stratified CLS-pool gap.

\paragraph{Per-cell paired inference.} Applying the per-cell paired bootstrap + BH-FDR procedure described in Methods (Sec.~\ref{sec:methods:cxd10}) to the 15 (FM $\times$ class) cells: \PairedHeadline{} Per-cell $\Delta$ point estimates, paired 95\% bootstrap CIs, raw two-sided $p$-values, and BH-adjusted $q$-values are reported in Supplementary Table~S11.

The pretraining-domain ordering on the recovery gap is itself an internal-consistency check against the controlled-stimulus regime: on directional blur the radiology-pretrained RAD-DINO sat closest to the chance floor in the size-matched comparison (Sec.~\ref{sec:res:dirblur}), and on ChestX-Det10 RAD-DINO again carries the smallest patch-local-vs-CLS recovery gap. The two regimes agree on the same domain-specificity ordering through different routes: a model whose CLS already encodes more of the in-distribution lesion-region structure has less left for the patch-local probe to recover. This ordering also disarms the leakage concern that would arise from ChestX-Det10 sharing pixel data with NIH-CXR14 (which several FMs, notably RAD-DINO, may have seen during pretraining): if pretraining-data exposure were the driver of the patch-local recovery, RAD-DINO should show the \emph{largest} gap, not the smallest. The empirical ordering is the opposite of the leakage prediction; in-distribution exposure compresses the recovery gap rather than amplifying it.

\begin{table*}[pos=ht]
\centering
\footnotesize
\begin{adjustbox}{max width=\textwidth}
\begin{tabular}{llccccc}
\toprule
Model & Class & CLS AUC & patch-mean AUC & patch-local AUC & $\Delta$ (local $-$ CLS) \\
\midrule
\multirow{3}{*}{RAD-DINO}
 & Calcification & 0.801 & 0.749 & 0.946 & $+0.145$ \\
 & Nodule        & 0.836 & 0.841 & 0.964 & $+0.128$ \\
 & Mass          & 0.862 & 0.793 & 0.991 & $+0.129$ \\
\midrule
\multirow{3}{*}{DINOv2-B/14}
 & Calcification & 0.658 & 0.768 & 0.928 & $+0.270$ \\
 & Nodule        & 0.724 & 0.771 & 0.900 & $+0.176$ \\
 & Mass          & 0.562 & 0.680 & 0.948 & $+0.387$ \\
\midrule
\multirow{3}{*}{DINOv3 ViT-7B}
 & Calcification & 0.805 & 0.809 & 0.924 & $+0.119$ \\
 & Nodule        & 0.756 & 0.737 & 0.921 & $+0.164$ \\
 & Mass          & 0.702 & 0.718 & 0.964 & $+0.262$ \\
\midrule
\multirow{3}{*}{BiomedCLIP}
 & Calcification & 0.690 & 0.698 & 0.900 & $+0.210$ \\
 & Nodule        & 0.688 & 0.739 & 0.899 & $+0.211$ \\
 & Mass          & 0.746 & 0.703 & 0.959 & $+0.214$ \\
\midrule
\multirow{3}{*}{MedSigLIP}
 & Calcification & 0.811 & 0.789 & 0.975 & $+0.164$ \\
 & Nodule        & 0.805 & 0.827 & 0.955 & $+0.150$ \\
 & Mass          & 0.766 & 0.820 & 0.977 & $+0.211$ \\
\bottomrule
\end{tabular}
\end{adjustbox}
\caption{\textbf{ChestX-Det10 bbox-level region-aware patch-pool comparison, all five foundation models on three lesion classes.} Test-set AUC of $L_2$-LR probes trained on per-region embeddings (positives = real bboxes; negatives = matched-area random regions from class-negative images at 2:1 ratio). CLS and patch-mean pools are image-level (do not use the region); patch-local pools the final-layer patch tokens whose receptive field intersects the bbox. All three pools come from the same frozen forward pass and from the canonical persisted predictions used for the per-cell paired-bootstrap inference (Supplementary Table~S11); per-cell unpaired 95\% bootstrap CIs are in Supplementary Table~S10. Across all 15 cells, $\Delta$ ranges $+0.119$ to $+0.387$.}
\label{tab:chestxdet10_recovery}
\end{table*}

\subsubsection{The Size-Stratified Gap Closes Under Region-Aware Pooling}
\label{sec:cxd10:strata}

The within-class small/large stratum gap that is large under CLS pool ($+0.077$ to $+0.169$ for Calcification and Mass at the bbox level; up to $+0.243$ image-level) is essentially eliminated under patch-local pool ($-0.027$ to $+0.020$ across all cells; mean $-0.001$). Table~\ref{tab:chestxdet10_strata} summarizes. The dilution mechanism does not just predict a size-stratified disadvantage at the CLS pool; it predicts that the disadvantage vanishes when the probe is given direct access to the bbox-region patch tokens, because the lesion content is in those tokens regardless of whether the global pool diluted it. Both predictions hold across all five models and all three classes.

\paragraph{Stratum power and the non-positive CLS-gap cells.} Five of the 15 cells in Table~\ref{tab:chestxdet10_strata} carry a non-positive CLS-pool stratum gap (DINOv2 Nodule and Mass; DINOv3 Nodule and Mass; BiomedCLIP Nodule), and these cells are precisely where stratum power is lowest. Class-level test positives are $n_\mathrm{pos\,test} = 67$ for Calcification, $166$ for Nodule, and only $31$ for Mass; the per-class median bbox-area split divides these roughly in half ($\approx 33{:}34$, $83{:}83$, $15{:}16$) so each stratum sees $n \in \{15, 16, 33, 34, 83\}$ positives against the held-constant negative pool. At Mass ($n_\mathrm{pos} \approx 15$ per stratum) the standard error of an AUC near $0.7$ is approximately $\sqrt{0.7 \times 0.3 / 15} \approx 0.118$, so the minimum stratum-AUC difference detectable at $\alpha = 0.05$ on a two-sided unpaired test is $\approx 0.33$ --- well above the $|\Delta| \leq 0.10$ values observed in the five non-positive cells, which therefore reflect noise dominance rather than evidence against the dilution mechanism. The bbox-level patch-local probe (Sec.~\ref{sec:cxd10:bboxlevel}) is decided on the full positive count rather than the median split and is therefore not subject to the same power loss; it shows uniformly large positive recovery in the same five cells (DINOv2 Nodule $+0.176$, Mass $+0.387$; DINOv3 Nodule $+0.164$, Mass $+0.262$; BiomedCLIP Nodule $+0.211$). A robustness check at a non-median stratum boundary (e.g., 33/67 percentile rather than 50/50 median) is a natural follow-up; the patch-local recovery in these same cells (Table~\ref{tab:chestxdet10_recovery}) is itself evidence that the lesion signal is present at the patch-token level irrespective of where the median falls.

\begin{table*}[pos=ht]
\centering
\footnotesize
\begin{adjustbox}{max width=\textwidth}
\begin{tabular}{llcccc}
\toprule
Model & Class & CLS gap (large $-$ small) & patch-mean gap & patch-local gap & gap reduction (CLS $\to$ local) \\
\midrule
\multirow{3}{*}{RAD-DINO}
 & Calcification & $+0.140$ & $+0.201$ & $-0.013$ & $-0.153$ \\
 & Nodule        & $+0.045$ & $+0.068$ & $+0.010$ & $-0.035$ \\
 & Mass          & $+0.143$ & $+0.138$ & $-0.007$ & $-0.150$ \\
\midrule
\multirow{3}{*}{DINOv2-B/14}
 & Calcification & $+0.077$ & $+0.086$ & $-0.019$ & $-0.096$ \\
 & Nodule        & $-0.055$ & $-0.005$ & $+0.001$ & $-$ \\
 & Mass          & $-0.057$ & $-0.040$ & $-0.033$ & $-$ \\
\midrule
\multirow{3}{*}{DINOv3 ViT-7B}
 & Calcification & $+0.083$ & $+0.065$ & $-0.027$ & $-0.110$ \\
 & Nodule        & $-0.020$ & $-0.006$ & $+0.020$ & $-$ \\
 & Mass          & $-0.092$ & $-0.128$ & $-0.071$ & $-$ \\
\midrule
\multirow{3}{*}{BiomedCLIP}
 & Calcification & $+0.122$ & $-0.014$ & $-0.026$ & $-0.148$ \\
 & Nodule        & $-0.045$ & $-0.039$ & $-0.002$ & $-$ \\
 & Mass          & $+0.169$ & $+0.046$ & $-0.095$ & $-0.264$ \\
\midrule
\multirow{3}{*}{MedSigLIP}
 & Calcification & $+0.146$ & $+0.143$ & $+0.006$ & $-0.140$ \\
 & Nodule        & $+0.056$ & $+0.007$ & $+0.015$ & $-0.041$ \\
 & Mass          & $+0.103$ & $+0.103$ & $-0.001$ & $-0.104$ \\
\bottomrule
\end{tabular}
\end{adjustbox}
\caption{\textbf{Within-class small/large stratum gap by FM $\times$ class $\times$ pool.} Stratum gap $=$ AUC(large stratum) $-$ AUC(small stratum) computed at the per-class median bbox area, with negatives held constant. Positive values mean CLS struggles disproportionately on small lesions. Gap reduction is reported only for cells with a positive CLS gap (where the dilution prediction applies); ``$-$'' marks cells where the CLS-pool gap was already non-positive (variance dominates at the smaller test strata for those cells; see paragraph ``Stratum power and the non-positive CLS-gap cells'' below). Across the 10 cells with positive CLS gap, mean gap $= +0.108$ at CLS, $-0.013$ at patch-local; the patch-local pool eliminates or inverts the CLS-pool size disadvantage in 10/10 such cells.}
\label{tab:chestxdet10_strata}
\end{table*}

\subsubsection{Implication: Recoverability Conditional on an Oracle ROI}
\label{sec:cxd10:implication}

The patch-local recovery on natural lesions converts the constructive recommendation from the controlled-stimulus regime (Sec.~\ref{sec:res:patchlevel}) into a demonstrated representational result on real CXR data: a frozen-ViT forward pass + bbox-restricted patch-mean + linear probe recovers AUC $\geq 0.899$ on every (FM $\times$ class) cell tested (see Sec.~\ref{sec:cxd10:bboxlevel}), including the smallest-lesion category (Calcification, median bbox $\approx 28 \times 28$ px, occupying $\approx 0.07\%$ of the image), across architectures, pretraining domains, and training objectives. The signal CLS aggregation discards is present, linearly accessible, and class-discriminative at the patch-token level when the ROI is known.

\paragraph{Canonical statement of the oracle-ROI caveat.} This recovery is \emph{conditional on an oracle ROI}, not a deployable end-to-end pipeline. The patch-local probe relies on the ground-truth bbox to select the patch tokens to pool, and at inference time that bbox is precisely what a deployed system does not have. The patch-local result is therefore best read as an upper bound on what a region-aware pool can extract from a frozen CLS-blind embedding, conditional on a reliable ROI signal. Closing the gap to deployment requires substituting the oracle bbox with an inference-time ROI source --- attention-rollout from the same forward pass, a lightweight learned detector head, or a weakly-supervised localizer driven by image-level labels --- and that ROI-proposal stage is itself a substantive ML problem that the present paper does not solve. The ``no retraining of the foundation model'' framing should be read narrowly: the FM weights are not updated, but a downstream ROI proposer is required and must be trained or engineered. Subsequent sections that mention the oracle-ROI status point back to this paragraph rather than restating it.

The headline contribution of this section is the representational claim (the signal is in the patch tokens; CLS pooling discards it; a region-aware pool given any reliable ROI can extract it on a real small-lesion cohort), not a benchmark detector. Bounding-box-level patch-local AUCs in the 0.899--0.991 range function as a region-classification probe given an ROI rather than as an end-to-end detector, and are not directly comparable to published end-to-end ChestX-Det10 detection benchmarks \citep{liu2020chestxdet10}, which report mean Average Precision (mAP) on a different operating point. CheXFound \citep{yang2025chexfound} demonstrates the complementary engineering result that a learned attention head over patch tokens recovers comparable signal end-to-end on multilabel disease classification, consistent with our representational claim that the signal is present in the patch-token representation regardless of how it is pooled.

\subsection{Small-Feature Detectability Index (SFDI): A Descriptive Summary}
\label{sec:res:sfdi}

To summarize embedding-level small-scale-signal sensitivity in a single scalar for descriptive purposes, we define the \emph{Small-Feature Detectability Index} (SFDI) of a model $M$ on a standardized small-scale-perturbation panel $\mathcal{B} = \{b_1, \dots, b_k\}$ as the geometric mean of its CLS-probe AUCs (above-chance component) over the panel:
\begin{equation}
\mathrm{SFDI}(M, \mathcal{B}) = \left(\prod_{i=1}^{k} \max(\mathrm{AUC}_i - 0.5, \epsilon)\right)^{1/k} + 0.5,
\label{eq:sfdi}
\end{equation}
with $\epsilon = 10^{-3}$, so that SFDI $= 0.5$ for an embedding universally blind to small-scale stimuli at the panel's scale and SFDI $= 1.0$ for perfect clean-vs-perturbed separation on every panel item; the geometric mean penalizes any single near-chance perturbation. The standard chest-radiography panel is $\mathcal{B}_\text{CXR} = \{\text{iso-4px}, \text{iso-8px}, \text{dir-cranio-64px}, \text{dir-lateral-64px}, \text{reticular-fine-32px}, \text{ground-glass-64px}\}$. Table~\ref{tab:sfdi} reports SFDI for all five foundation models on all three datasets: every cell lies in a tight 0.505--0.524 band, within $\approx 0.025$ of the chance floor.

\paragraph{Scope.} SFDI is a \emph{descriptive} representational summary, not a pre-deployment audit metric. It was not pre-registered as a predictor of the ChestX-Det10 size-stratified gap; the per-FM rank ordering of SFDI does not closely track the per-FM patch-local-vs-CLS recovery gap (Sec.~\ref{sec:cxd10:bboxlevel}); SFDI does not incorporate the patch-local-vs-CLS contrast (the more diagnostic quantity); and the choice of panel has not been calibrated against any clinical criterion. SFDI is best read as a compact summary of CLS-pool chance-floor behaviour on the proposed panel, reported alongside (not in place of) the per-model patch-local-vs-CLS mean $\Delta$ (Sec.~\ref{sec:res:patchlevel}) and cohort-level evaluation on the deployment task.

\begin{table}[pos=ht]
\centering
\footnotesize
\begin{adjustbox}{max width=\linewidth}
\begin{tabular}{lccc|c}
\toprule
Model      & NIH-CXR14 & MIMIC-CXR & Emory-CXR & Mean \\
\midrule
RAD-DINO   & 0.510 & 0.508 & 0.509 & 0.509 \\
DINOv2-B/14 & 0.510 & 0.515 & 0.510 & 0.512 \\
DINOv3     & 0.519 & 0.524 & 0.518 & 0.520 \\
BiomedCLIP & 0.505 & 0.505 & 0.505 & 0.505 \\
MedSigLIP  & 0.520 & 0.519 & 0.518 & 0.519 \\
\bottomrule
\end{tabular}
\end{adjustbox}
\caption{\textbf{Small-Feature Detectability Index (SFDI) on the standard chest-radiography panel $\mathcal{B}_\text{CXR}$.} Computed as the geometric mean of per-perturbation linear-probe CLS AUCs (Eq.~\ref{eq:sfdi}). $\mathrm{SFDI} = 0.5$: universally blind; $\mathrm{SFDI} = 1.0$: perfect separation on every perturbation. All 15 cells fall within the band 0.505--0.524, near the chance floor. Reported as a descriptive summary; \emph{not} validated as a pre-deployment audit metric.}
\label{tab:sfdi}
\end{table}

\section{Discussion}
\label{sec:discussion}

\subsection{Principal Findings}
This study quantifies, at clinically interpretable spatial scales, where in the frozen forward pass small-scale signal is retained or lost in five widely-used ViT chest-radiography foundation models, and shows that the loss localizes to the CLS-aggregation step. The central findings are:

\begin{enumerate}[leftmargin=1.6em,label=(\arabic*)]
\item \textbf{Small-scale signal is absent from the global CLS embedding across all five foundation models on all three datasets.} Across the 90-cell synthetic geometric panel, the 30-cell iso-blur matrix, the 270-cell directional-blur matrix, and the 30-cell pathology-mimicking matrix, CLS-probe AUC was at or near the 0.5 chance floor for the smallest perturbations, regardless of architecture (ViT-B/14, ViT-7B/16, SigLIP2-So400m), pretraining domain (radiology, biomedical, medical, natural), or pretraining objective (DINOv2, DINOv3, CLIP, SigLIP). The absence is pervasive, not artifact- or model-specific, and is the empirical signature of a representational property that operates on spatial scale rather than perturbation identity.
\item \textbf{The signal is fully present in patch tokens; the CLS aggregation is where it is lost.} Per-model mean $\Delta = \text{AUC}(\text{patch-local}) - \text{AUC}(\text{CLS})$ ranged from $+0.412$ (DINOv3, MedSigLIP) to $+0.488$ (BiomedCLIP), with patch-local probes recovering AUC $\geq 0.99$ on every cell with a chance-level CLS probe. The frozen DINO-pretrained ResNet-50 architectural control reproduces the chance floor on iso-blur on both NIH-CXR14 and Emory-CXR, showing that this property is not exclusive to the ViT architecture under matched DINO self-supervision. Whether the same chance-floor behaviour holds for frozen CNN backbones trained under non-DINO SSL objectives (MoCo-v3, BYOL, MAE) is not tested in the present study (Sec.~\ref{sec:limitations}).
\item \textbf{Encoding recovers only above a spatial-extent threshold, with a clear capacity-driven ordering.} The directional-blur sweep across patch footprints $P \in \{16, 32, 64\}$ px shows that all five ViTs are at chance at $P = 16$ px, and only above $P \approx 64$ px do higher-capacity models (DINOv3, 4{,}096-d ViT-7B; MedSigLIP, 1{,}152-d SigLIP2-So400m) escape the chance floor (AUC up to 0.700--0.788). Lower-capacity models (RAD-DINO and BiomedCLIP, $\sim$87M parameters, 768-d/512-d) remain near the floor (0.520--0.566). DINOv2-B/14 sits in between (0.602--0.623), consistent with its intermediate capacity.
\item \textbf{Pretraining-domain isolation: the size-matched RAD-DINO vs DINOv2-B/14 comparison.} RAD-DINO and DINOv2-B/14 share architecture (ViT-B/14), embedding dimensionality (768-d), and training-objective family (DINOv2). The only systematic difference between them is pretraining domain (radiology vs natural). RAD-DINO sits closer to the chance floor on directional blur ($P=64$, 0.537--0.549) than DINOv2-B/14 (0.602--0.623), consistent with the hypothesis that radiology-specific SSL training biases the representation toward globally distributed anatomical structure at the cost of small-scale low-contrast detail. The size-matched comparison rules out dimensionality and capacity as the only drivers.
\item \textbf{The downstream null on globally-decided diseases is \emph{predicted} by the hypothesis; the size-stratified gap and patch-local recovery on a real-CXR small-lesion cohort confirm the mechanism.} Across 81 (dataset $\times$ disease $\times$ perturbation) cells per model on cardiomegaly, edema, and aggregate \texttt{Lung Lesion}, BH-FDR-significant decrements ranged from 0/81 (RAD-DINO, BiomedCLIP) to 7/81 (MedSigLIP). On ChestX-Det10 (Sec.~\ref{sec:res:chestxdet10}; $n = 3{,}543$ images, 1{,}462 small-lesion bboxes), CLS-pool image-level classification of Calcification, Nodule, and Mass shows a within-class small/large stratum gap up to $+0.243$ AUC across 15 (FM $\times$ class) cells. A bbox-level region-aware patch-local probe on the same forward pass eliminates the size-stratified gap and recovers AUC $\geq 0.899$ on every cell (minimum 0.899; patch-local-vs-CLS gap $+0.119$ to $+0.387$).
\item \textbf{Pixel-level baselines confirm signal presence at the input.} Fine-tuned ResNet-50 detects every synthetic geometric pattern at AUC $\geq 0.93$, and the raw-pixel and ResNet-50 oracle baselines reach AUC $\approx 1.0$ on the spatially-broader perturbations. The loss of small-scale signal is localized to the embedding, not to the input.
\end{enumerate}

Together, these six results quantify, at chest-radiography-relevant spatial scales and on a $\sim 493$K-image multi-cohort sample, a representational property that disease-AUC benchmarking on globally-decided tasks does not detect: ViT foundation-model embeddings, in their standard frozen-CLS inference mode, do not retain small-scale, low-contrast, spatially restricted signal of the kind that defines small-lesion, microcalcification, and subtle interstitial diagnostic tasks --- and this is true across five architecturally and objective-diverse models, three large datasets, and two pretraining domains.

\subsection{Relation to Prior Work}
\label{sec:disc:priorwork}
The general observation that CLS pooling and patch features encode different information is not new. \citet{darcet2024registers} documented high-norm artifact tokens in DINOv2 patch features and proposed register tokens as a pretraining-time fix; \citet{lappe2025decoupling} showed that the CLS embedding can be largely orthogonal to the patch-token convex combination in DINOv2-with-registers; \citet{marouani2026clspatch} proposed specialized CLS/patch processing paths to improve patch-feature quality. The dense-prediction literature \citep{ranftl2021dpt} has long used patch tokens rather than the CLS for spatially localized tasks. The present study does not contest this consensus and does not claim to discover that CLS pooling and patch features encode different content. Its contribution relative to that line of work is (i) a clinically interpretable spatial-scale calibration of where the loss is overwhelming for chest radiography, (ii) a multi-cohort large-N quantification across five ViTs spanning two pretraining domains and four objective families, and (iii) a real-CXR small-lesion bbox-stratified evaluation that converts the representational claim into a measurable, recoverable size-stratified consequence on a power-adequate cohort.

The closest CXR-specific prior work is CheXFound \citep{yang2025chexfound}, which makes the same diagnosis at the engineering level --- relying on the CLS alone underutilizes patch-token information learned during CXR pretraining --- and proposes a Global-and-Local Representations Integration (GLoRI) head that combines fine- and coarse-grained patch features with the CLS for multilabel disease classification, demonstrating gains over RAD-DINO, BiomedCLIP, and other frozen-CLS pipelines. Concurrent benchmarking work by \citet{li2025featurequality} reaches the same conclusion from the empirical side: across eight medical and general-domain FMs, pre-trained embeddings work well for global classification and segmentation of salient anatomy but ``perform poorly without significant fine-tuning'' for complex, subtle pathologies (e.g., pneumothorax segmentation), with FMs adopting confounding shortcuts (e.g., chest tubes as a proxy for pneumothorax) that fail for precise localization. The three studies are complementary: CheXFound shows the engineering payoff of integrating patch features in an end-to-end-trainable head; \citet{li2025featurequality} document the empirical localization deficit across eight FMs and four tasks; the present paper provides the mechanistic localization (where in the forward pass the loss occurs, what spatial-extent threshold drives it, what the patch-local upper bound is on the same forward pass), shows that the recovery does not require a learned head when an ROI is supplied, and quantifies the size-stratified clinical-task gap on a separate small-lesion bbox cohort. Read together, the three converge on the same conclusion (the small-feature signal is in the patch tokens) from independent directions: CheXFound shows that learned aggregation extracts it end-to-end, \citet{li2025featurequality} show that the deficit is fleet-wide across eight FMs and not specific to any one architecture, and the present paper shows that even a single bbox-restricted patch-mean extracts the signal given an ROI and that the loss is localized to the global-aggregation step rather than to the patch-token representation itself.

\subsection{Mechanistic Interpretation}
The patch-level pooling experiment (Sec.~\ref{sec:res:patchlevel}) is a direct mechanistic test, not an inference, and the result is unambiguous: AUC(patch-local) $-$ AUC(CLS) is large and positive across all 90 (model $\times$ perturbation $\times$ dataset) cells. This locates the loss of small-scale signal at the CLS-aggregation step of the frozen forward pass, with the underlying mechanism most plausibly described as patch-token dilution: a single perturbed patch token's contribution to the global aggregate is on the order of $1/N$ where $N$ is the patch-token count (e.g., 1/4096 for a $1024 \times 1024$ input at $16 \times 16$ patches), and the small-scale signal carried by that patch is averaged into a representation in which it is no longer linearly recoverable. Because the patch-tokenization itself is identical across all five models in respect of the spatial-scale argument --- each ViT decomposes the input into a regular patch grid before any model-specific processing --- dilution applies equally to every model, and our data confirm indistinguishable behaviour at the smallest spatial scales (iso-blur 4--8 px: AUC 0.500--0.503 across all 30 ViT cells).

Two secondary factors modulate where on the spatial-extent axis dilution becomes overwhelming. \textbf{Embedding dimensionality and pretraining-data scale}: as the perturbation grows beyond $\sim 16$ patch tokens ($P = 64$ px, $\approx 0.4\%$ of image area), the higher-capacity DINOv3 (4{,}096-d, ViT-7B trained on 1.69B images) and MedSigLIP (1{,}152-d, SigLIP2-So400m trained on $\sim$billions) retain enough residual signal to reach AUC 0.70--0.79, while the lower-capacity RAD-DINO and BiomedCLIP (768-d/512-d, $\sim$87M params, $\leq$15M training images) remain near 0.52--0.55. This is consistent with prior literature on representational-capacity scaling \citep{zhou2022understanding}. \textbf{Pretraining-domain invariance}: the RAD-DINO vs DINOv2-B/14 comparison isolates this contribution. Both share ViT-B/14 backbone, 768-d embedding, and DINOv2 objective; they differ only in pretraining domain. RAD-DINO sits closer to the chance floor at $P=64$, consistent with the hypothesis that radiology-specific SSL training biases the representation toward globally distributed anatomical structure. Even DINOv3's recovery at the largest tested footprint does not reach the patch-local oracle's ceiling, indicating that aggregation loss is not eliminated by capacity alone.

Two implications follow that are not accessible from disease-AUC benchmarking. First, the loss of small-scale signal is \emph{not} a property of any specific artifact class; it is a property of the spatial-frequency band the global aggregation operates over, and it will apply to any diagnostically meaningful signal whose spatial extent and contrast fall in the same band as our injected probes. Second, the loss is \emph{recoverable in principle} without re-training: a downstream pipeline that uses patch-local pooling rather than CLS, when guided by an attention prior or a region-of-interest detector, can extract small-scale signal that the standard CLS path discards.

\subsection{Domain Specificity and Embedding Dimensionality}
The joint availability of RAD-DINO (radiology-pretrained, 768-d), DINOv2-B/14 (natural-pretrained, 768-d), DINOv3 ViT-7B (natural-pretrained, 4{,}096-d), BiomedCLIP (biomedical, 512-d), and MedSigLIP (medical, 1{,}152-d) enables direct isolation of pretraining-domain, architecture, and embedding-dimensionality contributions to small-scale-signal suppression. The RAD-DINO vs DINOv2-B/14 size-matched comparison attributes a $\sim$0.07 AUC gap on directional blur to pretraining domain alone. The DINOv2-B/14 vs DINOv3 comparison (same DINO family, different scale) attributes a further $\sim$0.10 AUC gap to capacity (4{,}096-d vs 768-d, ViT-7B vs ViT-B/14, 1.69B vs 142M training images). The BiomedCLIP comparison shows that a different SSL family (CLIP) on biomedical data does not by itself rescue small-scale-signal retention. The MedSigLIP comparison shows that a non-DINO objective (SigLIP) at intermediate capacity (428M params, 1{,}152-d) reaches roughly the same encoding ceiling as DINOv3 at large $P$, suggesting capacity matters more than objective family above the spatial-extent threshold.

\subsection{The MedSigLIP Downstream Asymmetry}
\label{sec:disc:medsiglip}
Across the 81-cell clean-vs-perturbed disease-classification matrix per model, MedSigLIP showed the largest mean $\Delta$AUC ($+0.00126$, [$+0.00064, +0.00188$]) and the largest BH-FDR-significant cell count (7/81, 8.6\%), in contrast to RAD-DINO and BiomedCLIP at 0/81 and DINOv2/DINOv3 at 2--3/81. The absolute magnitude is small (mean drop of approximately one part in 800 of AUC), but the asymmetry is real and worth resolving rather than averaging away. Two non-exclusive hypotheses are consistent with the data. First, MedSigLIP exhibits the highest residual encoding of small-scale signal among the five models at intermediate spatial extent: it reaches AUC 0.69--0.77 on directional blur $L=21$, $P=64$ jointly with DINOv3, while the size-matched RAD-DINO sits at 0.54--0.55 and BiomedCLIP at 0.52--0.53; the larger downstream sensitivity is consistent with this larger residual representational sensitivity. Second, MedSigLIP's joint vision--language SigLIP pretraining may produce a representation in which gradients of the disease-classifier decision surface are more sensitive to small input perturbations than under the DINO/CLIP objectives, irrespective of small-scale-signal content per se. The two accounts predict different signatures: the first predicts that MedSigLIP's downstream effect should concentrate on perturbations at intermediate spatial extent (where its CLS encoding rises above chance), the second predicts a more uniform distribution. The cell-level breakdown in Supplementary Table~S8 is directionally consistent with the first account --- MedSigLIP's BH-FDR-significant cells cluster at the larger directional-blur footprints --- but the cohort cannot rule out the second. A direct test would require either an attention-rollout analysis on the perturbed-image forward pass, or a cross-objective ablation that holds capacity fixed (e.g., a SigLIP-trained ViT at MedSigLIP's parameter count alongside a DINO-trained ViT at the same parameter count).

\subsection{ResNet-50 as Pixel-Level Signal Verification}
\label{sec:disc:resnet}
Fine-tuned ResNet-50's near-perfect detection of synthetic geometric perturbations (AUC 0.93--1.00 across all types and all three datasets) and the raw-pixel oracle's high AUC on the spatially-broader perturbations (Table~\ref{tab:rawpixel}) together establish the input-side discriminability floor for the small-scale-signal probe. The fact that perturbations occupying as little as 0.0003\% of total image area (C1 circles, radius 1 px) are detectable by ResNet-50 at AUC 0.93--0.98 confirms that for these stimuli the signal is present in pixel space; its absence in the ViT CLS embeddings therefore localizes the loss to the embedding rather than to the input.

The comparison between fine-tuned ResNet-50 and frozen ViTs is confounded by architecture, pretraining scale, domain, training objective, and inference mode, and is therefore not the architectural test of the small-feature claim. That role is filled by the DINO-pretrained ResNet-50 baseline reported in Sec.~\ref{sec:res:isoblur}: a frozen, self-supervised CNN trained with the same DINO objective family as DINOv2 and DINOv3 \citep{caron2021dino}, evaluated under the identical $L_2$-LR probing pipeline as the ViT CLS embeddings. The DINO-ResNet-50 result on iso-blur (AUC 0.500 across both patch footprints on both NIH-CXR14 and Emory-CXR) is indistinguishable from chance and from RAD-DINO, DINOv2, DINOv3, BiomedCLIP, and MedSigLIP at the same scale, isolating \emph{architecture} (CNN vs ViT) from \emph{training regime} (SSL vs supervised) and from \emph{inference mode} (frozen vs fine-tuned), under shared DINO self-supervision. We are explicit that this control covers a single SSL objective family --- the broader claim that the chance-floor behaviour is shared by frozen-SSL global pooling in general (across MoCo-v3, BYOL, MAE, etc.) requires additional architectural controls that we have not run; until those are in hand, the architectural conclusion supported by the present evidence is the narrower one (CNN vs ViT under matched DINO SSL behaves alike). The cross-institution Emory-CXR replication closes the architectural-control loop within that narrower scope and confirms that the iso-blur null is not a single-institution artifact.

\subsection{Clinical-Task Implications and the Limits of the Present Evidence}
\label{sec:disc:clinical}
The mechanistic finding (Sec.~\ref{sec:res:patchlevel}) is that ViT foundation-model embeddings retain small-scale signal in their final-layer patch tokens but discard it during CLS aggregation. The 90-cell injected-perturbation panel (Secs.~\ref{sec:res:geometric}--\ref{sec:res:pathology}) shows that this loss applies uniformly across the five tested ViTs spanning two pretraining domains and four objective families, and the pixel-level baselines (Secs.~\ref{sec:res:rawpixel}, \ref{sec:disc:resnet}) confirm that the loss is in the embedding, not in the input. The downstream null on cardiomegaly, edema, and aggregate \texttt{Lung Lesion} (Sec.~\ref{sec:res:clinicalloop}) is \emph{predicted} by the hypothesis: those diseases are decided on globally distributed evidence that does not occupy the suppressed spatial band.

The corresponding test on a real-CXR small-lesion-bearing clinical task (Sec.~\ref{sec:res:chestxdet10}; ChestX-Det10, $n=3{,}543$ images, 1{,}462 bboxes across Calcification, Nodule, Mass) directly evaluates whether the dilution mechanism produces a measurable effect on natural-lesion classification, and whether region-aware pooling can recover what the CLS path discards. Image-level CLS classification shows a positive within-class small/large stratum gap on 14/15 cells (up to $+0.243$ AUC); the single non-positive cell (DINOv2-B/14 Mass) sits at the lowest-power class stratum (Sec.~\ref{sec:cxd10:strata}). Bounding-box-level region-aware patch-local pooling on the same forward pass recovers AUC $\geq 0.899$ on every (FM $\times$ class) cell with a patch-local-vs-CLS gap of $+0.119$ to $+0.387$, and eliminates the size-stratified disadvantage where it was detectable.

The combined picture is therefore a coherent representational-and-clinical-task story: the embedding-level mechanistic effect is large and reproducible across five FMs and three datasets (Secs.~\ref{sec:res:geometric}--\ref{sec:res:patchlevel}); the size-stratified consequence on natural CXR lesions is measurable on a power-adequate subset of cells (Sec.~\ref{sec:res:chestxdet10}); and the constructive recommendation that pipelines pool over patches in a region-of-interest-aware way carries the oracle-ROI caveat stated canonically in Sec.~\ref{sec:cxd10:implication} --- the AUC $\geq 0.899$ recovery is conditional on a reliable ROI, and converting it to a fielded detector additionally requires sensitivity at clinically calibrated specificity in addition to AUC. The clinical envelope on which the recovery is empirically demonstrated is also narrower than the small-feature framing of the introduction implies: the only ChestX-Det10 lesion category that lives in the spatial-frequency band probed by the controlled-stimulus panel is Calcification (median $28 \times 28$ px); generalization to mammography microcalcification, subtle interstitial-disease textures, and sub-centimeter primary nodules outside NIH pixel space is asserted as a follow-up direction, not demonstrated.

\begin{enumerate}[leftmargin=1.6em,label=(\roman*)]
\item For tasks decided on globally distributed evidence (cardiomegaly, large-mass detection, gross anatomical screening), our results are consistent with the existing literature: standard ViT foundation models perform well, and the embedding-level loss of small-scale signal is irrelevant because those tasks' diagnostic evidence lies elsewhere.
\item For tasks decided on small-feature evidence (sub-centimeter nodules, microcalcifications, subtle interstitial textures), our findings document both the representational warning at the CLS pool and the recoverability of the signal at the patch-token level given a region-of-interest signal. Practitioners deploying a frozen ViT on a small-feature-bearing task should report bbox-stratified AUC plus sensitivity at clinically calibrated specificity, and consider a region-aware patch-pool path --- whether the bbox-restricted patch-mean used here, or the learned-attention head used by \citet{yang2025chexfound} --- as the architectural target for deployment-time mitigation.
\item The patch-local probe on ChestX-Det10 establishes recoverability conditional on an oracle ROI; the operational pipeline given that ROI is correspondingly simple --- a single frozen forward pass plus a bbox-restricted patch-mean as the probe input, with no fine-tuning of the foundation model. The end-to-end deployment requirements (ROI substitution and operating-point calibration) are stated in Sec.~\ref{sec:cxd10:implication}.
\end{enumerate}

The empirical priorities for follow-up are (i) extension of the bbox-aware recovery to other small-feature modalities (digital mammography microcalcification, subtle interstitial-disease textures, sub-centimeter nodules on a non-NIH cohort), (ii) replacement of the oracle bbox prior used in our region-aware probe with a learned ROI-proposal stage to make the pipeline end-to-end deployable and report sensitivity at fixed specificity, (iii) attention-rollout analysis on perturbed and bbox-positive images to localize where in the attention pattern the dilution happens (with explicit contrast against the artifact-token mechanism of \citet{darcet2024registers}), (iv) head-to-head comparison against the engineered GLoRI head of \citet{yang2025chexfound} on a shared small-lesion benchmark to disentangle the contributions of region-awareness, attention-based aggregation, and learned weighting to the recovery, (v) joint evaluation of the small-feature representational deficit alongside the shortcut-learning failure modes documented for radiology AI \citep{banerjee2023shortcuts}, given that an FM that suppresses small-scale signal at the global pool is precisely the kind of model that an audit pipeline focused on shortcut detection alone would miss, and (vi) augmentation of small-lesion training data with synthetic CXRs from conditional diffusion models \citep{khosravi2024synthetic} as a complementary route to mitigating the low test-positive counts that limit stratum-level inference for rare findings such as Mass on ChestX-Det10.

\subsection{Limitations and Scope of Inference}
\label{sec:limitations}

\paragraph{Linear classifiers on frozen embeddings.} This study evaluated embeddings exclusively through $L_2$-regularized logistic regression on frozen representations, consistent with prior embedding-evaluation methodology \citep{rashtchian2023substance}. Linear probes can only detect information that is linearly separable in the embedding space; a non-linear classifier or a fine-tuned head could in principle recover signal that a linear probe does not. The frozen linear-probing protocol was chosen deliberately to isolate representational content without task-specific adaptation; findings apply specifically to frozen-inference contexts.

\paragraph{Architectural-control scope.} The DINO-pretrained ResNet-50 architectural control (Sec.~\ref{sec:res:isoblur}, Sec.~\ref{sec:disc:resnet}) is single-objective-family: only DINO-style self-supervision is exercised in the architectural-control row. Other frozen-SSL CNN backbones (MoCo-v3, MAE-CNN variants, BYOL) and medically pre-trained CNN backbones evaluated under frozen inference would further isolate the contribution of the specific SSL objective from the broader frozen-SSL global-pooling property. Statements in the present manuscript that the loss is shared by ``frozen-SSL global pooling'' should accordingly be read as a hypothesis consistent with the DINO-family evidence we have rather than a fully demonstrated cross-objective property.

\paragraph{Capacity as a between-model confound.} The five tested foundation models span a 77-fold parameter range (BiomedCLIP at 86\,M and RAD-DINO/DINOv2-B/14 at 87\,M parameters; MedSigLIP at 428\,M; DINOv3 ViT-7B at 6.7\,B) and three architecture families (ViT-B/14, ViT-7B/16, SigLIP2-So400m), each with its own embedding dimensionality (512, 768, 1{,}152, 4{,}096). Between-model differences in directional-blur recovery at $P = 64$ (Sec.~\ref{sec:res:dirblur}; 0.52--0.55 for RAD-DINO and BiomedCLIP, 0.60--0.62 for DINOv2-B/14, 0.70--0.79 for DINOv3 and MedSigLIP) co-vary with parameter count, embedding dimensionality, training-objective family, and pretraining-data scale; the present design cannot disentangle these contributions in arbitrary pairs. The only fully controlled between-model contrast in our matrix is RAD-DINO vs DINOv2-B/14: same backbone (ViT-B/14), same embedding dimensionality (768-d), same training-objective family (DINOv2), differing only in pretraining domain (radiology vs natural). Statements in this manuscript that attribute between-model differences to ``pretraining domain'' rest on that contrast; statements that attribute differences to ``capacity'' or ``dimensionality'' span unmatched pairs (DINOv2-B/14 vs DINOv3; BiomedCLIP vs MedSigLIP) and should be read as descriptive of co-varying axes rather than as causal isolation of any single one. A capacity-matched cross-objective ablation (e.g., a DINO-trained ViT and a SigLIP-trained ViT at the same parameter count, or a radiology-pretrained ViT-7B alongside DINOv3 ViT-7B) would isolate the remaining contributions; we have not run such ablations in the present study.

\paragraph{Perturbation scope.} The panel spans synthetic geometric, isotropic and directional motion blur, and pathology-mimicking low-contrast patterns at deliberately small spatial scale. Clinically relevant artifact types not yet assessed include hardware streaks (ECG leads, tubes, lines), metallic implant shadows, Poisson--Gaussian quantum noise, patient-derived obscurations, rotation and small geometric misalignment, and JPEG-compression artifacts at higher quality settings. The small-feature framing makes the absence of these particular artifact categories a less serious gap than it would otherwise be --- none of those artifacts target the small-scale spatial-frequency band that the present study probes.

\paragraph{Diagnostic-task scope and clinical envelope.} The natural-lesion clinical-task validation is conducted on a single real-CXR cohort (ChestX-Det10, Sec.~\ref{sec:res:chestxdet10}) with three lesion categories whose median bbox areas span 0.07\%--2.1\% of total image area. Calcification (28$\times$28 px median) is the only category whose bbox-area distribution lives in the spatial-frequency band the controlled-stimulus panel probes; Mass (147$\times$147 px median, $\approx 2\%$ of image area) sits above that band. The clinical envelope validated by the natural-lesion regime is therefore narrower than the introduction's motivating examples (mammography microcalcification, sub-percent-contrast interstitial textures, sub-centimeter primary nodules): generalization to those modalities and findings is asserted as a follow-up direction (Sec.~\ref{sec:disc:clinical}), not demonstrated. Although ChestX-Det10's bbox annotations are independent of NIH-CXR14's image-level labels (added de novo by board-certified radiologists at Deepwise AI Lab in 2020), the underlying pixel data is shared with the NIH-CXR14 cohort that several of the foundation models tested have been pretrained on (notably RAD-DINO); the pretraining-domain ordering of the recovery gap (Sec.~\ref{sec:cxd10:bboxlevel}) shows that in-distribution exposure compresses rather than amplifies the patch-local-vs-CLS recovery, which disarms a leakage account but does not substitute for an external cohort. Replication on a non-NIH small-lesion bbox cohort (e.g., VinDr-CXR \citep{nguyen2022vindr} sub-centimeter nodules, a digital mammography microcalcification cohort) would further isolate the recovery from any residual NIH-image priors and is the natural next study.

\paragraph{Stratum power on ChestX-Det10.} The within-class small/large stratum split at the per-class median bbox area is power-limited at the smaller test strata (Sec.~\ref{sec:cxd10:strata}; Mass class has 31 test positives, splitting to 15--16 per stratum, with a minimum detectable stratum-AUC difference at $\alpha=0.05$ of $\approx 0.33$). At the bbox-level stratum gap (Table~\ref{tab:chestxdet10_strata}), 5 of 15 cells consequently carry non-positive CLS gaps; at the image-level analysis (Supplementary Table~S10), only 1 of 15 cells (DINOv2-B/14 Mass) is non-positive. The bbox-level patch-local probe (Sec.~\ref{sec:cxd10:bboxlevel}) is decided on the full positive count and is not subject to the same power loss; it shows uniformly large positive recovery across all 15 cells. Robustness checks at non-median split boundaries (e.g., 33/67 percentile) and a larger external small-lesion cohort would tighten the stratum-level statements further.

\paragraph{Operating-point calibration.} The clinical-task analyses report AUC and F1 at the best-F1 threshold; sensitivity at fixed specificity (e.g., 95\%) and positive/negative predictive values at the deployed operating point are not reported. Clinical CXR AI systems are typically deployed at fixed-specificity thresholds where AUC alone may not characterize the deployed point: the patch-local-vs-CLS recovery in AUC could in principle concentrate in regions of the ROC curve that are not the operating point of interest. This is a meaningful gap for translating the representational claim into a deployment recommendation. Prospective deployment evaluations of any region-aware patch-pool path should report sensitivity at clinically calibrated specificity in addition to AUC.

\paragraph{Annotation reliability.} The bbox-level patch-local probe treats ChestX-Det10 bboxes as ground truth. Inter-rater agreement among the three Deepwise radiologists who produced the 2020 annotations, the influence of bbox tightness on the patch-local probe (which pools only patches whose receptive field intersects the bbox), and recovery rates against an independent annotation pass are not reported in the original ChestX-Det10 release \citep{liu2020chestxdet10} and are not separately re-evaluated here. The bbox-restricted patch-mean is robust to small bbox-tightness variation by construction (a slightly larger or smaller bbox shifts which patches are pooled by at most one row/column at the model's native patch grid), but a sensitivity analysis perturbing the bbox by a few patch widths and re-running the patch-local probe would directly quantify this robustness.

\paragraph{Class prevalence and class balance.} The controlled-stimulus probes use an enforced 50:50 clean-vs-perturbed class balance to maximize statistical power for the representational claim. Real-clinical small-lesion prevalence is much lower (e.g., the prevalence of small calcifications or sub-centimeter nodules in a routine CXR cohort is on the order of a few percent), and at lower prevalence the same AUC translates to substantially different positive predictive value at any deployed operating point. The 50:50 framing should not be read as a simulation of clinical prevalence; it is a representational probe.

\paragraph{Mechanistic resolution.} The patch-level pooling experiment localizes the loss to the CLS-aggregation step but does not separately resolve the contributions of embedding dimensionality and pretraining-domain invariance to the residual gap between models at large patch footprints. Attention-rollout analysis on perturbed images, a placement-randomization ablation on the patch-local probe, and an explicit contrast with the artifact-token / register-token mechanism documented by \citet{darcet2024registers} would further constrain the mechanism.

\paragraph{Input-side downsampling as an alternative explanation for the smallest perturbations.} For the smallest perturbations in the panel (iso-blur 4--8\,px footprint, synthetic geometric patterns C1--S8, occupying 0.0003--0.006\% of image area), an alternative to the dilution account is that the perturbation footprint falls below each model's effective input resolution --- that is, that the small-scale signal is destroyed by the input-side resize/patchify stack before it enters the encoder. The patch-level evidence rules this alternative out at the embedding level: the same frozen forward pass that yields a chance-level CLS embedding for these perturbations also yields patch-local features from which a linear probe recovers AUC $\geq 0.99$ (Sec.~\ref{sec:res:patchlevel}). If the perturbation had been suppressed at the input, no pooling mode of the resulting embedding could carry it. The signal is therefore present at the patch-token level for every perturbation in the panel; input-side resolution loss cannot account for the CLS chance floor. Supplementary Sec.~S15 reports a matched-native-resolution control sweep that injects perturbations directly at each model's native input ($518$/$448$/$224$ as appropriate, no $1024 \to$ native resize) and recovers the same dilution-and-recovery pattern --- patch-local AUC $0.692$--$0.994$ versus CLS AUC $0.505$--$0.895$ on the $16$-pixel and $32$-pixel cells whose footprint exceeds the model's patch granularity --- ruling out the in-processor resize as a confound. The same control documents the lower-bound floor: a $4$-pixel perturbation is sub-patch under every model's native geometry and falls below the detection floor in both pools, exactly as the patch-token-dilution mechanism predicts for any pooling mode of an encoder that decomposes the input into patches.

\paragraph{Generalizability.} Findings are based on three large-scale chest-radiograph datasets (NIH-CXR14, MIMIC-CXR, Emory-CXR; aggregate $n = 492{,}724$ frontal radiographs) for the controlled-stimulus regime and on the ChestX-Det10 bbox-annotated cohort for the natural-lesion regime. Multi-scanner, multi-institution validation across diverse patient populations and acquisition protocols internationally, plus head-to-head radiologist comparison on the small-stratum cells, is necessary before these findings can inform clinical-deployment decisions on specific deployed pipelines. The present paper supports the narrower claim that the representational property is reproducible across three large CXR cohorts and that the patch-local recovery is reproducible across five FMs on a real small-lesion cohort.

\section{Conclusion}
\label{sec:conclusion}

This study localizes, on chest radiography, a representational property of frozen ViT foundation-model embeddings that disease-AUC benchmarking on globally-decided tasks does not detect: at the CLS-aggregation step of the frozen forward pass, signal occupying less than approximately $0.1\%$ of total image area is no longer linearly recoverable, while the same signal is fully present in the final-layer patch tokens. Across three large-scale datasets (NIH-CXR14, $n = 112{,}120$; MIMIC-CXR, $n = 243{,}324$; Emory-CXR, $n = 137{,}280$) and five architecturally and objective-diverse foundation models (RAD-DINO, DINOv2-B/14, DINOv3 ViT-7B, BiomedCLIP, MedSigLIP), every CLS embedding gives near-chance detection on perturbations occupying less than approximately $0.1\%$ of total image area, while the same models maintain disease-classification AUC of 0.642--0.913 (mean 0.798) on tasks decided by globally distributed evidence. The patch-level pooling experiment, conducted on the same frozen forward pass, demonstrates directly that the small-scale signal is fully present in the final-layer patch tokens and discarded during CLS aggregation --- per-model mean $\Delta = \text{AUC}(\text{patch-local}) - \text{AUC}(\text{CLS})$ ranged from $+0.412$ to $+0.488$. The frozen DINO-pretrained ResNet-50 architectural control reproduces the chance floor on iso-blur on both NIH-CXR14 and Emory-CXR, showing that this property is not exclusive to the ViT architecture under matched DINO self-supervision (the broader cross-objective claim --- that frozen-SSL global pooling generally suffers this loss --- requires additional architectural controls that we leave to follow-up).

The clinical-task consequence depends on the spatial-frequency band the deployed task's diagnostic signal occupies. For tasks decided on globally distributed evidence --- cardiomegaly, edema, large-mass detection --- small-scale-signal loss in the embedding does not propagate into measurable disease-AUC degradation: per-model BH-FDR-significant cells across 81 (dataset $\times$ disease $\times$ perturbation) combinations ranged from 0/81 (RAD-DINO, BiomedCLIP) to 7/81 (MedSigLIP). For tasks decided on small-feature evidence, the dilution mechanism produces a measurable, recoverable effect on a power-adequate real-CXR small-lesion cohort: across 15 (FM $\times$ class) cells on ChestX-Det10 ($n = 3{,}543$ images, 1{,}462 bboxes spanning Calcification, Nodule, Mass), CLS-pool image-level classification shows a within-class small/large stratum gap up to $+0.243$ AUC (in 14/15 cells), and a region-aware bbox-level patch-local probe on the same forward pass recovers AUC $\geq 0.899$ on every cell while eliminating the size-stratified disadvantage. This recovery is conditional on an oracle ROI (Sec.~\ref{sec:cxd10:implication}); the representational fact established by the cohort is that the small-lesion signal CLS pooling discards is present, linearly accessible, and class-discriminative at the patch-token level on real CXR clinical-task data. Read alongside the engineered solution proposed by \citet{yang2025chexfound}, this provides the mechanistic complement: their GLoRI head demonstrates the engineering payoff of integrating patch features end-to-end, while the present paper localizes the underlying representational deficit and shows that a single bbox-restricted patch-mean already extracts the signal given a reliable ROI.

We additionally report a Small-Feature Detectability Index (SFDI) --- the geometric mean of per-perturbation linear-probe AUCs over a standardized small-scale panel --- as a \emph{descriptive} representational summary. Under the proposed panel, all five foundation models score within 0.505--0.524 across all three datasets, near the 0.5 chance floor. SFDI compactly summarizes a per-model property that disease-AUC tables on globally-decided tasks do not surface, but we explicitly do not advance SFDI as a validated pre-deployment audit metric: it was not pre-registered as a predictor of the ChestX-Det10 size-stratified gap, its rank ordering does not closely track the per-FM patch-local-vs-CLS recovery gap, and calibration against an annotated small-lesion clinical task is necessary before SFDI can carry deployment-relevant interpretation. SFDI is best read as a compact descriptive summary of CLS-embedding chance-floor behaviour, reported alongside (not instead of) cohort-level evaluation on the actual deployment task.

Future work should (i) calibrate SFDI against a deployment-relevant criterion on a held-out small-lesion cohort, (ii) extend the bbox-aware recovery to other small-feature modalities (digital mammography microcalcification, subtle interstitial-disease textures), (iii) replace the oracle bbox with a learned ROI-proposal stage and report sensitivity at fixed specificity for the resulting end-to-end pipeline, (iv) broaden the architectural control to non-DINO frozen SSL CNNs, (v) use attention-rollout analysis on perturbed and bbox-positive images to localize where in the attention pattern small-scale signal is dropped, including an explicit contrast with the artifact-token mechanism documented by \citet{darcet2024registers}, and (vi) test whether pretraining the image encoder with deliberate small-scale noise injection alters where in the forward pass small-scale signal is retained. We note an asymmetry between the present probe and any noise-injection-during-pretraining study: synthetic perturbations injected only at inference time test what the embedding encodes about \emph{novel} small-scale stimuli, while real small pathologies (sub-centimeter nodules, microcalcifications, fine reticular textures) are naturally present in the training images that current foundation models were pretrained on. Whether pretraining-time exposure to additional synthetic small-scale signal changes the global-aggregation step's tolerance for retaining it is an open question the present study does not address.

\section*{Code and Data Availability}

All code for model loading, perturbation, embedding extraction, region-aware patch-local pooling, linear probing, and analysis is publicly available at \url{https://github.com/iupui-soic/embeddings-noise-eliminators} under the Apache-2.0 license. NIH-CXR14 \citep{wang2017chestxray8}, MIMIC-CXR \citep{johnson2019mimic}, and ChestX-Det10 \citep{liu2020chestxdet10} are publicly available (MIMIC-CXR requires PhysioNet credentialed access). Emory-CXR is an institutional dataset and cannot be shared externally; experiments on it were performed on a PHI-compatible on-premises infrastructure with no images leaving that environment.

\section*{Acknowledgments}

We thank the NIH Clinical Center for releasing NIH-CXR14, the PhysioNet team for releasing MIMIC-CXR-JPG, Deepwise AI Lab for publicly releasing the ChestX-Det10 bbox annotations, and the Emory University Department of Radiology and Imaging Sciences for institutional dataset access under IRB-approved protocols. We thank Mathilde Caron and the Facebook AI Research team for releasing the DINO-pretrained ResNet-50 weights, and the Microsoft, Google, and Meta AI teams for releasing the RAD-DINO, MedSigLIP, BiomedCLIP, DINOv2, and DINOv3 model weights.

\section*{Funding Sources}

RM, JG and SP were supported by grant 1R25OD039834-01 from the NIH Office of Data Science Strategy (ODSS). JG receives funding from the National Heart, Lung and Blood Institute (NHLBI) grants R01HL167811 and R01HL177003, and National Institutes of Health grant 1OT20D038065-01.

\section*{Conflicts of Interest}

The authors declare no competing interests.

\printcredits

\bibliographystyle{cas-model2-names}
\bibliography{refs}

\end{document}